\PassOptionsToPackage{table}{xcolor}
\documentclass[11pt, a4paper, goog]{google}

\usepackage[authoryear, sort&compress, round]{natbib}
\usepackage{longtable}
\usepackage{array}
\usepackage{float}
\usepackage{flafter}
\usepackage{placeins}
\usepackage{subcaption}
\usepackage{tikz}
\usepackage{pgfplots}
\usepackage{wrapfig}
\usepackage[skins,breakable]{tcolorbox}
\usepackage{xurl}
\graphicspath{{figures/}}
\sethlcolor{yellow}

\usetikzlibrary{positioning,arrows.meta}
\pgfplotsset{compat=1.18}

\definecolor{Ink}{HTML}{1B2A41}
\definecolor{Brand}{HTML}{078282}
\definecolor{Soft}{HTML}{F6F3EC}
\definecolor{Rule}{HTML}{D8D2C8}
\definecolor{Accent}{HTML}{C2622D}
\definecolor{Accent2}{HTML}{55606E}

\definecolor{lightblue}{RGB}{220,235,250}
\definecolor{RQTint}{HTML}{FFF8E5}
\definecolor{RQInk}{HTML}{765300}
\definecolor{HypTint}{HTML}{F1E8F0}
\definecolor{HypInk}{HTML}{4B0947}
\tcbset{
  calloutbox/.style={
    fonttitle=\bfseries\small,
    coltitle=white,
    enhanced,
    breakable,
    attach boxed title to top left={xshift=2.5mm,yshift=-2.5mm},
    left=2mm, right=2mm, top=1.5mm, bottom=1.5mm,
    width=\linewidth,
    arc=3.5mm
  }
}
\tcbset{
  takeawaysbox/.style={
    calloutbox,
    title=Takeaway,
    colback=HypTint,
    colframe=HypInk,
    colbacktitle=HypInk,
    boxed title style={rounded corners, size=small, colframe=HypInk, colback=HypInk}
  },
  hypothesisbox/.style={
    calloutbox,
    title=Hypothesis,
    colback=HypTint,
    colframe=HypInk,
    colbacktitle=HypInk,
    boxed title style={rounded corners, size=small, colframe=HypInk, colback=HypInk}
  },
  rqbox/.style={
    calloutbox,
    title=Research question,
    colback=RQTint,
    colframe=RQInk,
    colbacktitle=RQInk,
    boxed title style={rounded corners, size=small, colframe=RQInk, colback=RQInk}
  }
}

\definecolor{PHred}{HTML}{C81E1E}
\definecolor{PHblue}{HTML}{1D4ED8}
\definecolor{PHgreen}{HTML}{047857}

\providecommand{\comment}[1]{}
\renewcommand{\comment}[1]{{\color{PHblue}\textbf{[Review:}~#1\textbf{]}}}

\definecolor{TentativeInk}{HTML}{640D5F}
\newif\iftentativehighlight
\tentativehighlighttrue

\definecolor{Unclear}{HTML}{FFE84D}

\newif\ifrecoveredhighlight
\recoveredhighlighttrue

\definecolor{RecovInk}{HTML}{1F6F43}
\definecolor{RecovBG}{HTML}{DCF1E2}
\definecolor{RecovRule}{HTML}{2F855A}

\newcommand{\pass}[1]{pass@#1}
\newcommand{\acc}{acc@1}
\newcommand{\Ddiv}{\mathcal{D}_{\mathrm{div}}}
\newcommand{\Dsim}{\mathcal{D}_{\mathrm{sim}}}
\newcommand{\cand}{\mathcal{C}}
\newcommand{\verifier}{\mathcal{V}}
\newcommand{\fingerprint}{\phi}
\newcolumntype{Y}{>{\raggedright\arraybackslash}X}

\pgfplotsset{
  paper/.style={
    axis lines=left,
    tick align=outside,
    ymajorgrids=true,
    grid style={dashed, Rule!90},
    tick label style={font=\footnotesize, color=Ink},
    label style={font=\footnotesize, color=Ink},
    title style={font=\small\bfseries, color=Ink, yshift=2pt},
    legend style={draw=none, fill=none, font=\footnotesize},
    every axis plot/.append style={draw opacity=1, fill opacity=0.92},
  },
  every node near coord/.append style={font=\scriptsize, color=Ink},
}

\usepackage{etoolbox}
\AtBeginEnvironment{tabularx}{\setlength{\parskip}{0pt}}
\AtBeginEnvironment{tabular}{\setlength{\parskip}{0pt}}
\AtBeginEnvironment{longtable}{\setlength{\parskip}{0pt}}



\newif\iftwocolumnlayout
\newcommand{\sidefigwidth}{1.65in}
\twocolumnlayoutfalse
\iftwocolumnlayout
  \newcommand{\sidefig}[3][]{\begin{figure}[t]\centering #3\end{figure}}
\else
  \newcommand{\sidefig}[3][]{\begin{wrapfigure}[#1]{r}{\dimexpr#2+0.15in\relax}\centering
    \captionsetup{font=small,justification=raggedright,singlelinecheck=false}#3\end{wrapfigure}}
\fi

\tcbset{rqbox/.append style={unbreakable}, takeawaysbox/.append style={unbreakable}}


\uselogo{}

\title{Selecting Diverse SFT Traces Improves Post-RL Generalization}
\hypersetup{pdftitle={Selecting Diverse SFT Traces Improves Post-RL Generalization},
  pdfauthor={Dylan Zhang, Mingyuan Wu, Jinning Li}}

\correspondingauthor{Dylan Zhang, \href{mailto:shizhuo2@illinois.edu}{shizhuo2@illinois.edu}}

\author[1]{Dylan Zhang}
\author[2]{Mingyuan Wu}
\author[2]{Jinning Li}

\affil[1]{University of Illinois Urbana-Champaign, work done at Google}
\affil[2]{Google}

\begin{abstract}
Verified solutions are not equally useful for preparing reasoning models for reinforcement learning (RL). We present a comprehensive study of route diversity, the variation in the sequences of reasoning steps in supervised fine-tuning (SFT) data, and propose a lightweight, rule-based fingerprint to select for it. From one pool at one budget, with matched training recipes and checkpoints, selecting diverse rather than similar routes improves post-RL problem coverage across puzzles and mathematics, including on problems harder than those seen in either training stage. In synthetic experiments, route-diverse SFT improves OLMo3-7B's \pass{8} by 16.9 points on environments held out from SFT. In a single-model condition, where one model writes every candidate, diverse selection gains up to 6.2 points of mean \pass{8} across 10 mathematics benchmarks. Pre-RL diagnostics suggest why: diverse SFT can produce both successful and failed attempts on more prompts despite slightly lower mean accuracy, giving group-relative RL more prompts with a learning signal. On 3 open-source corpora, our CPU-only selector, without model calls, outperforms more expensive alternatives in every comparison of mean post-RL performance. These results identify reasoning-route diversity as a practical criterion for selecting SFT data that better prepares models for RL.
\end{abstract}

\begin{document}

\maketitle

\section{Introduction}

\begin{figure*}[!t]
\centering
\begin{subfigure}[t]{1.8in}
\centering
\includegraphics[width=\linewidth]{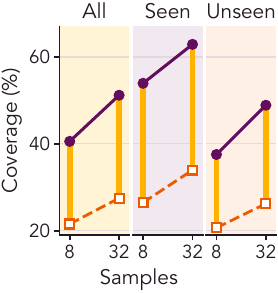}
\caption{Environment split}
\end{subfigure}\hfill
\begin{subfigure}[t]{1.60in}
\centering
\includegraphics[width=\linewidth]{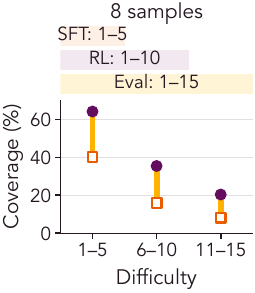}
\caption{Difficulty bands}
\end{subfigure}\hfill
\begin{subfigure}[t]{1.8in}
\centering
\includegraphics[width=\linewidth]{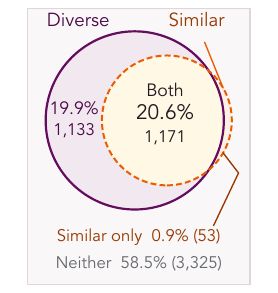}
\caption{Solved-set overlap}
\label{fig:rlve-olmo-coverage-expansion}
\end{subfigure}
\caption{OLMo3-7B on RLVE after the same RL. Diverse (purple circles)
solves more problems than Similar (orange open squares) on environments seen and
unseen in SFT and in every difficulty band. It solves almost all of what Similar
solves, plus 1{,}133 questions that Similar misses, while Similar solves 53 that
Diverse misses (c).
\textbf{(a,b)}~Gold marks the gap.
\textbf{(c)}~Solved sets at eight attempts, with gold marking the shared set.
Single runs. The generation cap is 16{,}384 tokens for (b) and the Seen/Unseen
8-sample points in (a), and 32{,}768 for the All curve, the 32-sample points,
and (c).}
\label{fig:rlve-olmo-performance}
\end{figure*}

The choice of supervised examples can shape a reasoning model long after
supervised training ends. A common post-training recipe first applies
supervised fine-tuning (SFT) on verified solutions, then reinforcement learning
with verifiable rewards (RLVR) on the model's own attempts
\citep{deepseekr1,kimik15,qwen3,seed15thinking,magistral,mimo,phi4reasoning,bercovich2025llamanemotron,lambert2024tulu3,olmo3,deepseekmath,yang2024qwen25math}.
Because RL samples from the policy that SFT produces, the starting
distribution influences which successful attempts it can discover within a
finite rollout budget
\citep{yue2025rlboundary,kim2025rlvsdistill,interplay2025,pear2026}.
Many recent systems keep pre-RL SFT lightweight
\citep{deepseekr1,qwen3,kimik15,glm45,meta2025llama4},
and studies caution that too much SFT can limit subsequent learning and
generalization
\citep{kang2025quagmires,jin2025rlheals,liu2026rejuvenation,li2026gettingready,sftmemorizes2025}.
The question is therefore not only how much supervised data to use, but
\emph{which verified solutions best prepare the model for the RL stage
that follows}.

Reasoning-data pipelines and self-training methods generate candidate
solutions, retain those that pass verification, and select a subset for
training
\citep{deepseekr1,qwen3,seed15thinking,bercovich2025llamanemotron,olmo3,star,yuan2023rft,guan2025rstarmath}.
Released reasoning corpora likewise provide pools of teacher-generated
solutions
\citep{guha2025openthoughts,muennighoff2025s1,ye2025limo}.
Selection commonly considers readability, length, reward scores, or
per-problem quotas
\citep{deepseekr1,wen2025lightr1,kimik15,grattafiori2024llama3,mimo,magistral};
other studies compare teachers and response counts
\citep{phi4reasoning,guha2025openthoughts,liu2025acereason11},
or emphasize instruction coverage, individual-trace quality, and student fit
\citep{lima2023,tulu2023,instag2023,deita2024,car2024,grape2025,bids2025,li2025naturalthoughts}.
These criteria do not directly characterize whether the retained solutions
provide different ways of reasoning or repeatedly demonstrate the same one.

We study this distinction through \emph{route diversity}.
A route is the sequence of reasoning steps taken by a verified solution;
route diversity describes how much the retained routes differ, including
among solutions to the same problem.
Two accepted solutions can reach the same answer through different
decompositions, explorations, and checks.
This distinction matters when reasoning is viewed as search over sequences
of steps: sampling varied chains improves inference-time reasoning
\citep{selfconsistency,diversityofthought2023,hao2023rap},
and training on varied search traces can improve the model's reasoning
\citep{mixeddistillation2023,gandhi2024sos}.
Our hypothesis is that practicing more varied routes can put correct
attempts within sampling reach on more problems, giving subsequent RL more
opportunities to learn.

Prior work provides important evidence for this hypothesis.
\citet{yuan2023rft} improve mathematical reasoning by adding distinct correct
solutions per problem, increasing data volume together with variety.
\citet{ju2025rpd} allocate a fixed demonstration budget to divergent solutions
for fewer problems and find that the advantage persists after RL.
Other studies change the teacher, SFT objective, timing of reasoning
supervision, or behaviors taught before RL
\citep{kim2025rlvsdistill,pear2026,frontloading2025,cen2025behaviorinjection,wang2026curiosft},
and show that SFT accuracy alone is an unreliable measure of readiness
for RL \citep{kang2025quagmires,li2026gettingready}.
Complementing work on which compositional experiences training must provide
\citep{kong2026modules}, we focus on the selection decision within an existing
reasoning-data pipeline:

\begin{tcolorbox}[rqbox]
At a fixed SFT demonstration budget, does selecting more varied reasoning
routes from the same candidate pool improve post-RL problem coverage,
and what distinguishes the resulting starting policies?
\end{tcolorbox}

We make two contributions: a comprehensive, controlled study that combines
teacher-source sweeps with direct route-selection experiments, and a simple,
scalable selection method built on a rule-based fingerprint.
Teacher count provides a coarse proxy for solution variety:
at fixed demonstration counts, multi-teacher SFT improves post-RL coverage
on synthetic puzzles
\citep{zeng2026rlve,stojanovski2025reasoninggym,chen2025enigmata},
out-of-distribution OMEGA problems \citep{sun2025omega},
and held-out mathematics (Section~\ref{sec:modelcount}).
We then compare route-diverse and route-similar selections from one pool
at one budget, matching the student initialization, training recipes,
evaluation protocol, and checkpoint step.
The fingerprint approximates procedural differences by summarizing
reasoning steps, their ordering, and path statistics
\citep{minegishi2025topology,xiong2025mapping,shahariar2025hierarchical}.
Selecting solutions that are spread out or concentrated in this
representation produces contrasting SFT datasets without changing the
training objective.

Route-diverse selection improves post-RL \emph{coverage}: the fraction of
held-out problems solved in at least one of a fixed number of attempts.
On RLVE, route-diverse SFT improves OLMo3-7B's \pass{8} by 16.9 percentage
points on environments held out from SFT, even though both conditions
subsequently receive RL on those environments.
The advantage extends to problems harder than those used in either
training stage and also appears with Qwen3 students
(Section~\ref{sec:topology}).
The solved sets show substantial expansion rather than merely an exchange
of successes: the diverse model solves 1{,}133 questions that the similar
model misses, versus 53 in the opposite direction
(Figure~\ref{fig:rlve-olmo-performance}\subref{fig:rlve-olmo-coverage-expansion}).

The benefit does not require multiple teachers.
In the \emph{single-model condition}, Qwen3-4B-Thinking-2507 writes every
candidate, and route-diverse selection from that one pool still improves
mean \pass{8} across 10 math benchmarks by
3.39 to 6.17 points at three selection budgets
(Section~\ref{sec:single-teacher}).

Pre-RL diagnostics suggest why this distinction matters.
With binary outcome rewards, rollout groups whose attempts all succeed
or all fail have zero group-relative advantage; mixed outcomes supply
the outcome-based learning signal \citep{deepseekmath,dapo2025,noprompt2026}.
Mean accuracy does not capture how frequently such groups occur across
prompts. On 64 mathematics training prompts, the route-diverse OLMo3-7B
checkpoint produces mixed outcomes on 54.7\% of prompts, compared with
46.9\% for the route-similar checkpoint, despite slightly lower mean
accuracy (Section~\ref{sec:analysis}).
This observation is consistent with route-diverse SFT providing a broader
distribution of learning opportunities before RL begins, rather than
simply a more accurate initialization.

Finally, the selection method we propose is useful beyond controlled
comparisons. Its rule-based, text-only fingerprint enables selection without model calls,
additional generation, or gradients, and runs on CPUs over candidate pools
exceeding two million solutions.
Applied to OpenThoughts3, INTELLECT-3, and Nemotron-Cascade~2
\citep{guha2025openthoughts,intellect3,nemotroncascade2},
the method beats random selection, the topology baseline (a simpler rule on the
same fingerprints), and gradient-diversity, embedding, and lexical selection
in every comparison of mean post-RL accuracy and \pass{8}
(Section~\ref{sec:realdata}).
This intervention acts before RL, complementing methods that filter or
reshape zero-variance groups \citep{dapo2025,noprompt2026},
maintain rollout diversity
\citep{chen2025passktraining,hu2025diver,wang2025forkingtokens},
or select prompts by reward variance and learnability
\citep{jiang2025vcrl,wu2026hive}.
Our study identifies route diversity as a practical selection criterion
alongside correctness and task coverage, and our selection method makes it
cheap to apply. Preparing a model for RL requires
attention not only to which problems its demonstrations solve, but also
to the variety of reasoning routes those demonstrations provide.

\section[Teacher Count as a Proxy for Route Diversity]{Teacher Count as a Proxy for Route Diversity}
\label{sec:modelcount}
\begin{tcolorbox}[takeawaysbox]
All else held equal, sourcing reasoning traces from a diverse teacher
pool beats a single-teacher counterpart in post-RL performance, even (1) when
the RL domain differs from the SFT domain and (2) when evaluating on
out-of-distribution benchmarks.
\end{tcolorbox}

\paragraph{Protocol.}
Every comparison in this paper follows one protocol. The two
conditions share the student model, the prompt pool, the SFT trajectory budget,
the group-relative RL recipe \citep{deepseekmath}, and the evaluation. They
differ only in \textbf{which solutions the SFT data contains}. Students are Qwen3 base checkpoints and OLMo3-7B, and candidate
generators are open reasoning models (Appendix~\ref{app:rosters}). Both conditions in every post-RL comparison
are read at the same RL step. Main results are
post-RL, and the initialization diagnostic of
Section~\ref{sec:analysis} is measured before RL.

\paragraph{Experiment Set-up.}
In this section, the SFT solutions come from one teacher or from
several teachers at the same trajectory budget. The SFT data are synthetic
puzzles with verifiable answers, with prompts from the 16-environment or the
399-environment subset of RLVE \citep{zeng2026rlve}. After SFT, each comparison
pairs one RL environment with its evaluation sets:
(i)~RL on Enigmata and evaluation on held-out Enigmata puzzles
\citep{chen2025enigmata}, (ii)~RL on the training set of OMEGA, a mathematics benchmark, and
evaluation on its out-of-distribution set, which has explorative, compositional
and transformative splits \citep{sun2025omega}, (iii)~RL on reasoning-gym tasks
\citep{stojanovski2025reasoninggym} and evaluation on OMEGA's out-of-distribution
compositional problems and on AIME 2024, AIME 2025, MATH-500 and Minerva, and
(iv)~RL on the DAPO-Math-17k mathematics set \citep{dapo2025} and evaluation on
the same four mathematics benchmarks. RL trains outside the SFT puzzles in every pair: on
another puzzle suite in (i), on reasoning-gym tasks in (iii), and on mathematics
in (ii) and (iv). In (iii), neither SFT nor RL trains on OMEGA or on the four
mathematics benchmarks.
Appendix~\ref{app:construction}
gives the protocols and the intermediate teacher counts.

\begin{figure*}[!t]
\centering
\begin{subfigure}[t]{1.7in}
\centering
\includegraphics[width=1.78in]{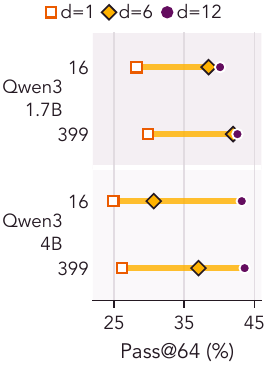}
\caption{Enigmata, 1, 6, and 12 teachers.}
\label{fig:mot-route}
\end{subfigure}\hfill
\begin{subfigure}[t]{1.7in}
\centering
\includegraphics[width=1.78in]{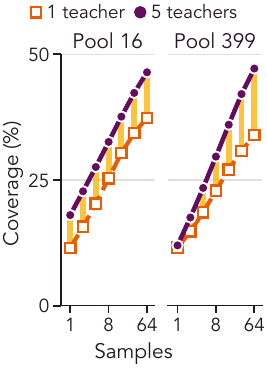}
\caption{RL on OMEGA.}
\label{fig:mot-ood-gain}
\end{subfigure}\hfill
\begin{subfigure}[t]{1.75in}
\centering
\includegraphics[width=1.80in]{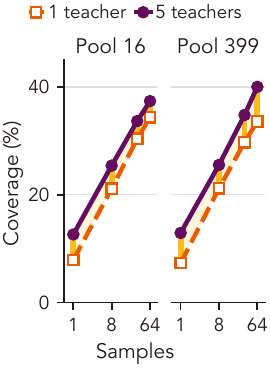}
\caption{RL on reasoning-gym}
\label{fig:response-diversity-omega-transfer}
\end{subfigure}
\captionsetup{justification=raggedright,singlelinecheck=false}
\caption{At a fixed SFT trajectory budget, more teacher sources give
higher post-RL coverage. \textbf{(a)}~Held-out Enigmata \pass{64}
after RL on Enigmata, for Qwen3-1.7B and Qwen3-4B students and SFT pools of 16
and 399 environments, where $d$ is the number of teachers. \textbf{(b, c)}~Qwen3-1.7B on OMEGA's
out-of-distribution compositional problems after RL on OMEGA's training set (b) and reasoning-gym (c).}
\label{fig:mot-route-and-source}
\end{figure*}

\begin{figure*}[!t]
\centering
\begin{subfigure}[t]{1.1in}
\centering
\includegraphics[width=\linewidth]{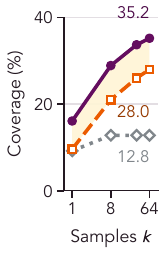}
\caption{1 and 12 teachers.}
\label{fig:mot-enigmata-decomp}
\end{subfigure}\hfill
\begin{subfigure}[t]{1.1in}
\centering
\includegraphics[width=\linewidth]{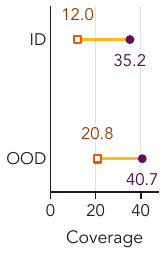}
\caption{Multi-vs-14B.}
\label{fig:mot-teacher-route}
\end{subfigure}\hfill
\begin{subfigure}[t]{3.19in}
\centering
\includegraphics[width=3.19in]{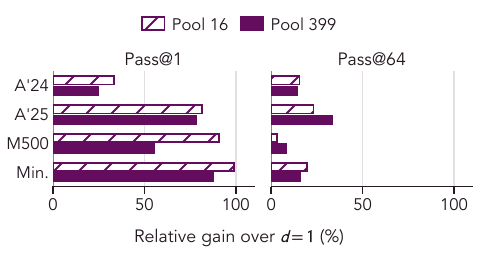}
\caption{Qwen3-4B mathematics, 1 or 12 teachers.}
\label{fig:teacher-source-math-qwen4b}
\end{subfigure}
\caption{At a fixed SFT trajectory budget, twelve teachers beat one
after RL on Enigmata (a) and after RL on DAPO-Math-17k (c).
\textbf{(a)}~Enigmata coverage of Qwen3-1.7B with twelve solutions per prompt
from twelve teachers (purple circles) or from one teacher (orange open
squares). Gray diamonds mark RL without SFT. \textbf{(b)}~Enigmata \pass{64} of
Qwen3-1.7B after RL for the verified twelve-teacher recipe (purple circles) and the
single-Qwen3-14B-teacher recipe (orange open squares), on in-domain (ID)
and out-of-domain (OOD) problems. \textbf{(c)}~Relative gain of
twelve teachers over one ($d{=}1$) on AIME 2024 (A'24), AIME 2025 (A'25),
MATH-500 (M500) and Minerva (Min.), hatched for the 16-environment pool and
solid for the 399-environment pool.}
\label{fig:mot-route-source}
\end{figure*}

\paragraph{(1) RL on a domain different from SFT.}
The Enigmata grid crosses one, six or twelve teachers with the two
environment pools (Figure~\ref{fig:mot-route}). For Qwen3-4B-Base, twelve
teachers add about 18 points of \pass{64} over one on held-out
Enigmata puzzles at either pool size, while
enlarging the pool from 16 to 399 RLVE environments adds up to 6.4 points at a fixed
teacher count and under half a point at twelve teachers. In a
Qwen3-1.7B comparison with twelve trajectories per prompt in both conditions, drawing them
from twelve teachers improves Enigmata coverage over drawing all twelve from one
(Figure~\ref{fig:mot-enigmata-decomp}). The verified twelve-teacher recipe
also beats the single-Qwen3-14B-teacher recipe, a
comparison of complete recipes described in Appendix~\ref{app:construction}
(Figure~\ref{fig:mot-teacher-route}).
Twelve teachers also lead when RL moves from puzzles to mathematics. We
take the Qwen3-4B-Base checkpoints after one-teacher and
twelve-teacher SFT on RLVE puzzles, train them with RL on DAPO-Math-17k, and
evaluate both at RL step 200. Twelve teacher sources then beat one
on AIME 2024, AIME 2025, MATH-500, and Minerva, in both environment pools
at \pass{1} and \pass{64} (Figure~\ref{fig:teacher-source-math-qwen4b}). On MATH-500 in the 16-environment pool, \pass{1} rises from
$34.14\%$ to $65.08\%$. Across the seven sampling budgets from
\pass{1} to \pass{64}, twelve teachers lead in 54 of the 56 benchmark, pool and
budget cells (Appendix~\ref{app:math-qwen4b-sources}). For Qwen3-1.7B
students trained by SFT with one to five teachers and then by RL on
DAPO-Math-17k, the multi-teacher conditions beat the one-teacher condition in
all 32 comparisons at \pass{1} and in 27 of 32 at \pass{64}
(Appendix~\ref{app:dapo}).
The lead also appears when RL runs on reasoning-gym tasks, which differ from
the RLVE puzzles used for SFT. After that RL stage, five teachers beat one on
OMEGA's out-of-distribution problems and on the four mathematics benchmarks
(Figure~\ref{fig:response-diversity-omega-transfer} and item (2) below).

\paragraph{(2) Out-of-distribution evaluation.}
OMEGA's out-of-distribution problems combine or transform skills beyond
OMEGA's training distribution. On the compositional split, Qwen3-1.7B
fine-tuned on solutions from five teachers beats the same student fine-tuned on
one teacher's solutions, at every reported sampling budget and in both
environment pools, in two sweeps with different RL training domains. The first runs RL on
OMEGA's training set (Figure~\ref{fig:mot-ood-gain}, complete sweep in Appendix
Figure~\ref{fig:app-mot-ood-full}). The second runs RL on
reasoning-gym tasks, so neither SFT nor RL trains on OMEGA
(Figure~\ref{fig:response-diversity-omega-transfer}). In the second sweep,
every multi-teacher condition from two to five teachers exceeds the one-teacher
condition at RL step 350 (Appendix Figure~\ref{fig:app-response-diversity-omega-full}).

The checkpoints behind
Figure~\ref{fig:response-diversity-omega-transfer}, after RL on reasoning-gym
tasks, are also evaluated on AIME 2024, AIME 2025, MATH-500 and Minerva, and
neither SFT nor RL trains on these benchmarks. Five teachers beat one on mathematics in every
benchmark and pool comparison at \pass{1} and \pass{64} (Appendix
Figure~\ref{fig:response-diversity-math-transfer}). The complete
sweep, including two reversals on Minerva with three and four teachers, appears
in Appendix Figure~\ref{fig:app-response-diversity-math-full}. Because teacher count is a proxy for route diversity,
Section~\ref{sec:topology} selects routes directly from one pool at one budget,
and there the route-diverse set leads even when one teacher writes every
candidate (Section~\ref{sec:single-teacher}).




\section[Selecting Verified Solutions for Route Diversity]{Selecting Verified Solutions for Route Diversity}
\label{sec:topology}

\begin{tcolorbox}[takeawaysbox]
(1) Reasoning diversity can be measured by topology and can be selected for directly.

(2) From one pool, at the same budget and with
the same RL, route-diverse SFT data reproduce the post-RL coverage advantage.

(3) Single-model condition: even from a single teacher, route-diverse selection leads to benefits.
\end{tcolorbox}

\subsection{Selecting verified routes}
\label{sec:measure}
\begin{figure*}[t]
\centering
\includegraphics[width=6in]{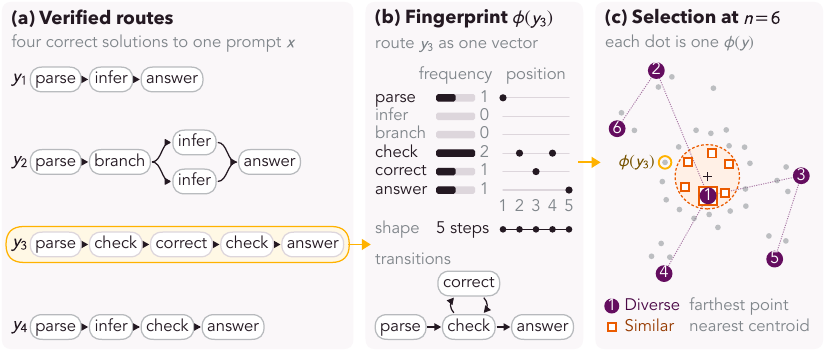}
\caption{Topology-based selection. \textbf{(a)}~Verified solutions to
one prompt $x$ as sequences of step events. \textbf{(b)}~Route $y_3$ (gold) as a
fingerprint of event frequencies, positions, shape, and transitions.
\textbf{(c)}~Selection in fingerprint space and at the same budget
$n=6$, nearest-centroid selection (orange) stays inside the dashed circle and
farthest-point selection (purple, numbered in order) spreads out.}
\label{fig:topodiv}
\end{figure*}

A route is the sequence of steps a verified solution takes from the
problem to the answer, such as a case split in mathematics, a move over the
board in Sokoban, or a rewrite of the program state in program simulation. Its
\emph{topology} is the structure of that path once wording, formatting, and
teacher identity are set aside, and two correct solutions to one problem can
have very different topologies (Figure~\ref{fig:topodiv}a).
This view of reasoning as paths and graphs of steps
follows prior work
\citep{treeofthoughts,besta2024got,ning2024skeleton,minegishi2025topology,xiong2025mapping,tan2025shape,shahariar2025hierarchical}.
We programmatically parse the topology of each verified solution $y$ as a
fingerprint $\fingerprint(y)$, a fixed-length vector describing that structure
(Figure~\ref{fig:topodiv}b). The fingerprint is simple to compute
and scales to whole candidate pools. It comes from step annotations where a
domain provides them and from the trace text otherwise. Read from the text, it
needs only fixed rules that label the steps and a fixed random projection that
shortens the vector, with no model calls, no new generation and no gradients.
Fingerprinting and selection run on CPUs over released pools of more than two
million solutions (Appendix~\ref{app:realdata-baselines}). Fingerprint
distances approximate differences in procedure and can also reflect wording.
Appendix~\ref{app:method} describes the fingerprint construction
and each domain's step vocabulary.

From one candidate pool we select two datasets of the same target size. For
the diverse set $\Ddiv$ we cluster the fingerprints, give each cluster a
size-proportional budget, and inside each cluster repeatedly add the candidate
farthest from those already chosen, a coreset construction \citep{coreset2018}.
Nearest-centroid selection builds the similar set $\Dsim$ from one dense region
(Figure~\ref{fig:topodiv}c). Every such comparison uses this procedure
(Appendix~\ref{app:method}).

\subsection{Settings and evaluation}
We run the selection on RLVE and evaluate every setting after RL by
sampled coverage. RLVE environments each provide a generator, a difficulty parameter, and
a rule-based verifier \citep{zeng2026rlve,stojanovski2025reasoninggym}, so we
choose which environments enter SFT and RL, hold some out of SFT, and evaluate
above the difficulties either stage used.

\paragraph{RLVE evaluation splits.}
The fixed held-out set spans RLVE environments and difficulties
(Appendix Table~\ref{tab:config}). Some questions have a programmatic reference answer,
and environment verifiers score the rest. A difficulty split separates a gain
inside the difficulties the training stages used (1 to 10) from a gain on harder
extrapolation problems (11 to 15). The Qwen3 runs report the in-range problems
with $\pass{32}$ and the extrapolation problems with $\pass{64}$. Both OLMo3-7B conditions select from one SFT environment set, and an
environment split marks its 63 evaluation environments as Seen and the 321 held
out from SFT as Unseen. The shared RL pool spans all 384 environments, so this split asks whether
the advantage reaches beyond the SFT task pool after the same RL. The OLMo3-7B
run reports it with $\pass{8}$ and $\pass{32}$ on the same prompts for both
conditions.
\subsection{RLVE: generalization across difficulty and environments}
\label{sec:results}
\newcommand{\singleteacherfigure}{%
\sidefig[13]{\sidefigwidth}{%
\includegraphics[width=\sidefigwidth]{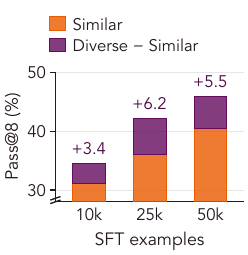}
\caption{One-teacher route selection. Orange tops: Similar; stack tops:
Diverse. Labels: point gain. Axis starts at 28\%.}
\label{fig:single-teacher}
}}
\iftwocolumnlayout\singleteacherfigure\fi
On RLVE \citep{zeng2026rlve}, we select SFT routes
from a shared pool and apply the same GRPO recipe to each pair. SFT covers
difficulty 1 to 5, RL extends through difficulty 10,
and evaluation runs to difficulty 15 on the same environments beyond training stages (Table~\ref{tab:config}).

For OLMo3-7B, at \pass{8} the diverse
condition leads by 16.9 points of coverage on environments held out from SFT
(Figure~\ref{fig:rlve-olmo-performance}). The margin is positive on both SFT-seen and SFT-unseen environments and increases
over the reported sampling budgets. Split by generator difficulty, the diverse
model leads in every band. The advantage persists across both splits: on problems harder than
either stage trained on, and on task families held out from SFT and included in
RL, so the difference set by the SFT selection survives a shared RL stage that
trained on both. The coverage advantage expands the solved set
(Figure~\ref{fig:rlve-olmo-performance}\subref{fig:rlve-olmo-coverage-expansion}). Diverse retains $95.67\%$ of Similar's solved questions and solves
1{,}133 that Similar misses, while Similar uniquely solves 53.

At both Qwen3 model sizes and both selection budgets, 50{,}000 and
200{,}000 SFT rows, Diverse leads on every reported metric, sampled \pass{1}
(Appendix Figure~\ref{fig:rlve-qwen-budget}) and sampled coverage, including on
extrapolation problems above the difficulty used in either SFT or RL.

The diverse selection also leads on held-out OMEGA mathematics
(Appendix~\ref{sec:sftcompound}) and on Sokoban. Program simulation
instead compares corpora from different generators, and the multi-model
corpus leads the single-model corpus. In Sokoban and program simulation each route can be replayed or
executed, and the gap grows with the number of samples
(Appendix~\ref{sec:olmo-explicit-routes}).

\subsection[Analysis: Mixed Rewards Before RL] 
{Analysis: Mixed Rewards Before RL}
\label{sec:analysis}

\paragraph{Why mixed rewards matter.}
With binary rewards, a group of $G$ independent rollouts on a prompt
$x$ with per-rollout success probability $p(x)$ is mixed with probability
\(
  P(\mathrm{mixed}\mid x)=1-p(x)^G-(1-p(x))^G,
\)
and a group whose rewards all agree yields zero group-relative advantage
\citep{deepseekmath,noprompt2026}. A prompt with no correct solution
within sampling reach has $p(x)$ near zero, so its group almost always fails
together and yields no update. Bringing one within reach makes a mixed group
possible. Mean solve rate averages $p(x)$ over prompts, so two policies with the
same accuracy can give RL different amounts of signal. If route-diverse SFT puts
a correct solution within sampling reach on more problems, that difference
should be visible before RL starts.

\begin{figure*}[!t]
\centering
\begin{minipage}{\linewidth}
\centering\small
{\color[HTML]{640D5F}$\bullet$}~Diverse\quad
{\color[HTML]{EB5B00}$\square$}~Similar\quad
{\color[HTML]{80868B}$\diamond$}~Pre-SFT base\quad
\raisebox{.25ex}{\textcolor[HTML]{FFB200}{\rule{1em}{2pt}}}~Favorable gap
\end{minipage}
\par\smallskip
\begin{subfigure}[t]{1.85in}
\centering
\includegraphics[width=\linewidth]{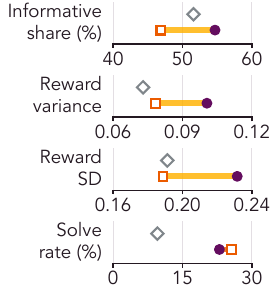}
\caption{Mathematics.}
\label{fig:rewardvar-init}
\end{subfigure}\hfill
\begin{subfigure}[t]{1.75in}
\centering
\includegraphics[width=\linewidth]{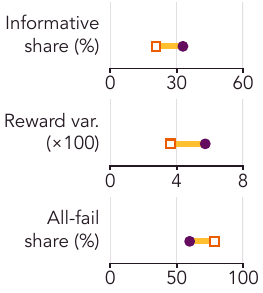}
\caption{RLVE: 8 samples.}
\label{fig:rlve-olmo-signal-8}
\end{subfigure}\hfill
\begin{subfigure}[t]{1.75in}
\centering
\includegraphics[width=\linewidth]{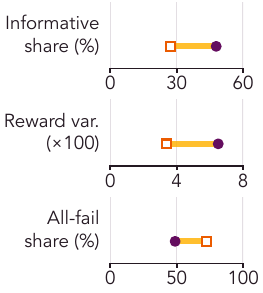}
\caption{RLVE: 32 samples.}
\label{fig:rlve-olmo-signal-32}
\end{subfigure}
\caption{OLMo3-7B reward-signal diagnostics before RL.}
\label{fig:olmo-reward-diagnostics}
\label{fig:rlve-olmo-signal}
\vspace{-3mm}
\end{figure*}

\newcommand{\answerdiversityfigure}{%
\sidefig[11]{\sidefigwidth}{%
\includegraphics[width=\sidefigwidth]{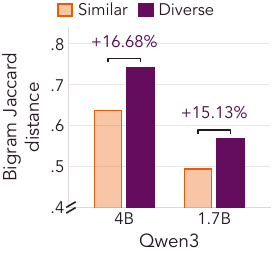}
\caption{Answer diversity of correct Qwen3 completions on RLVE after
the same RL.}
\label{fig:rlve-answer-diversity}
}}
\iftwocolumnlayout\else\answerdiversityfigure\fi
\paragraph{Mixed rewards before RL.}
We sample OLMo3-7B at the end of SFT on the Dolci-Think
diverse and similar 100{,}000-row selections (Appendix~\ref{app:dolci-checkpoints}),
before RL, eight times at temperature 1.0 on 64 mathematics prompts drawn from
the Dolci-RL-Zero-Mix prompts their RL trains on. The route-diverse checkpoint
has mixed rewards on $54.7\%$ of these prompts, against $46.9\%$
for the route-similar checkpoint and $51.6\%$ for the pre-SFT base,
at a slightly lower mean solve rate
(Figure~\ref{fig:rewardvar-init}). The two selections move this share
in opposite directions from the base. On held-out RLVE questions,
the diverse checkpoint of the RLVE OLMo3-7B pair in
Figure~\ref{fig:rlve-olmo-performance}, also before RL, likewise has more
mixed-outcome and fewer all-fail prompts at both budgets
(Figure~\ref{fig:rlve-olmo-signal}).

\iftwocolumnlayout\answerdiversityfigure\fi
\paragraph{Answer diversity after RL.}
On RLVE, after the same RL, the correct completions of the diverse
Qwen3 checkpoints of Section~\ref{sec:results} also vary more in wording
(Figure~\ref{fig:rlve-answer-diversity}). Their mean bigram Jaccard distance is
$16.68\%$ higher for Qwen3-4B and $15.13\%$ higher for Qwen3-1.7B than that of
the similar checkpoints. Appendix~\ref{app:answer-diversity} gives the
measurement protocol and longer-prefix comparison.

\paragraph{Discussion.}
\label{sec:discussion}
A post-training pipeline can verify far more solutions than its SFT
budget allows it to train on, so it has to keep a subset, and our results show
that this choice changes what the same RL can reach. At a fixed budget, keeping solutions whose routes differ
gives the same RL a better starting point, and selecting them from a pool the
pipeline already has needs no new generation.

\iftwocolumnlayout\else\singleteacherfigure\fi
\paragraph{Single-model condition.}
\label{sec:single-teacher}
The benefit does not depend on mixing
teachers. When one model, Qwen3-4B-Thinking-2507,
writes every candidate solution to Dolci-Think prompts \citep{olmo3} and both
sets are selected from that one pool at one budget, route-diverse SFT data
leads route-similar data after the same GRPO at all three SFT sizes, by 3.39
to 6.17 points of mean \pass{8} over ten competition-mathematics benchmarks
(Figure~\ref{fig:single-teacher}, configuration in Appendix
Table~\ref{tab:config}). The Qwen3-4B-Base student is
evaluated at the same RL step for both conditions within each SFT budget:
50 for 10k examples, and 30 for 25k and 50k.

\paragraph{Coverage gap by sampling budget and difficulty.}
The same account explains why the post-RL coverage gap widens with
the sampling budget: extra attempts recover more problems when more have a
correct solution within reach. In the illustrative model of
Appendix~\ref{app:model} the gap widens up to a finite budget, then narrows as
both policies approach saturation. The coverage gap peaks at
intermediate difficulty for Qwen3-4B-Base
(Appendix Figure~\ref{fig:app-rlve-perdiff-coverage}, accuracy by difficulty in
Appendix Figure~\ref{fig:app-rlve-perdiff-accuracy}), on the easiest band for OLMo3-7B
(Figure~\ref{fig:rlve-olmo-performance}), and at the smallest Qwen capacity on
OMEGA (Appendix Figure~\ref{fig:omega-models}). The account
predicts this pattern: the gap should be largest on problems near the edge of
what each starting model solves reliably.

\Needspace{5\baselineskip}
\section[Route Selection on Released Reasoning Corpora]{Route Selection on Released Reasoning Corpora}
\label{sec:realdata}
\begin{tcolorbox}[takeawaysbox]
1)The effect of SFT reasoning diversity holds in real-world corpus. 2)Our selection approach is cheap and scalable. It works on real-world datasets, outperforming more expensive baselines.  
\end{tcolorbox}

We now apply the selection to released reasoning corpora. For
OpenThoughts3 \citep{guha2025openthoughts}, INTELLECT-3 \citep{intellect3} and
Nemotron-Cascade 2 \citep{nemotroncascade2} we use the
solutions those corpora release directly. Both conditions are selected from
one pool at one budget, and an OLMo3-7B student receives the same
SFT and the same RL on a mixture of mathematics, code, instruction following,
and science. Appendix~\ref{app:dolci-checkpoints} describes the
Dolci-Think \citep{olmo3} selections used for the pre-RL diagnostic of
Section~\ref{sec:analysis}. Evaluation covers competition mathematics, the three OMEGA splits,
science, puzzle and instruction-following benchmarks
(Appendix~\ref{app:realdata-benchmarks}).

\begin{figure*}[t]
\centering
\includegraphics[width=5.95in]{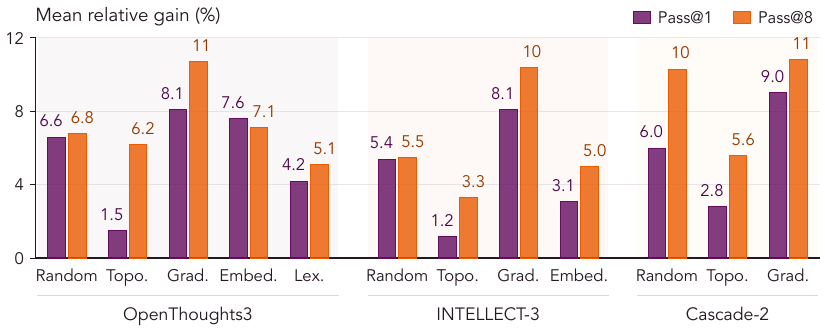}
\caption{Relative gain of \textbf{our selection over each selection baseline} in mean
score, with both conditions evaluated at the final RL checkpoint, step 64.
FPS :farthest-point selection.
Topo., Grad., Embed. and Lex. are the topology (our fingerprints with a
simpler rule), gradient-diversity, embedding and lexical baselines of
Appendix~\ref{app:baselines}.}
\label{fig:realdata-baseline-gains}
\end{figure*}

Our selection (Section~\ref{sec:measure}) beats random selection, the
topology baseline, and farthest-point selection on gradient,
embedding and lexical features in every comparison,
with relative gains in mean score from 1.2\% to 10.8\% at \pass{1} and
\pass{8} (Figure~\ref{fig:realdata-baseline-gains}).

Against the similar selection from the same pool, the diverse selection leads
on every mathematics benchmark in all three corpora, by 4.9 to 18.5 points of
average accuracy, and it also leads on all three OMEGA splits and on
GPQA-Diamond (Appendix~\ref{app:realdata-per-benchmark},
Appendix Figure~\ref{fig:realdata-corpora-intellect3}).

It also costs far less. On a
pool of about 2.1 million solutions it takes about three hours on one CPU
node and no GPU time, while the gradient-diversity and embedding baselines pass every
candidate through a 7B or 8B model and need 64 to 232 GPU-hours
(Appendix~\ref{app:selection-cost}). Appendix~\ref{app:realdata-baselines} gives the
selection procedures, and Appendix~\ref{app:realdata-relative-benchmarks} gives the
gains on each benchmark.

\section{Related Work}

\paragraph{Preparing a model for RL.}
The starting policy bounds what RL can reinforce
\citep{yue2025rlboundary,interplay2025}, and SFT response diversity predicts
post-RL performance
better than accuracy \citep{li2026gettingready,kang2025quagmires}. Prior work varies data timing
\citep{frontloading2025}, the SFT loss \citep{pear2026}, exploratory behaviors
\citep{cen2025behaviorinjection,wang2026curiosft}, reasoning primitives
\citep{yao2025tailor}, and the teacher \citep{kim2025rlvsdistill}.
\citet{kong2026modules} study how training on reasoning traces
builds reusable modules that support compositional generalization.
Comparisons of the two stages find that SFT memorizes more and RL generalizes
better \citep{sftmemorizes2025}, that RL lowers output diversity
\citep{kirk2024rlhf}, and that limiting how far SFT moves the policy preserves
more of its generality \citep{psft2025}.
\paragraph{Reasoning-data selection and structure.}
Data work curates small sets \citep{lima2023}, selects by
instruction diversity \citep{instag2023,deita2024,car2024}, trace quality
\citep{li2025naturalthoughts} or model fit \citep{grape2025,bids2025}, mixes
tasks \citep{sanh2021t0,selfinstruct,wizardlm,tulu2023} and isolates semantic
breadth \citep{diversification2025}. Closest to us, distinct paths per problem
raise post-SFT accuracy \citep{yuan2023rft}, and \citet{ju2025rpd} keep
divergent solutions for fewer problems at the same number of demonstrations,
with the advantage kept after RL. Their method calls a language
model on every candidate solution, which is costly at our pool sizes, whereas
route selection reads only the trace text. We
build on self-training \citep{star,yuan2023rft,gulcehre2023rest,singh2024restem},
coverage selection \citep{coreset2018,dpp2012}, reasoning graphs
\citep{minegishi2025topology,xiong2025mapping,tan2025shape,shahariar2025hierarchical}
and path search \citep{selfconsistency,treeofthoughts}, and complement measures \citep{friedman2023vendi,tevet2021evaluating,zhao2024measure}.
Appendix~\ref{app:related} covers methods that act on the reward signal inside
group-relative RL.

\section{Conclusion}
Verified solutions are not interchangeable as preparation for RL.
In our comprehensive study, at a fixed demonstration budget, under matched
training recipes and with evaluation at matched checkpoints,
route-diverse selection from the same pool improves post-RL coverage,
including on problems harder than either stage trained on.
The benefit persists in the single-model condition.
On three released corpora, our proposed CPU-only selector beats random
selection and the topology, gradient-diversity, embedding, and lexical
baselines in every comparison of mean post-RL performance, without model
calls or additional generation.

Pre-RL diagnostics suggest why accuracy alone can mislead:
the route-diverse OLMo3-7B checkpoint produces mixed rewards on
more prompts despite slightly lower mean accuracy, consistent with
giving group-relative RL more learning opportunities.
Appendix~\ref{sec:limitations} discusses limitations and future work.
Together, these results identify route diversity as a practical
criterion for choosing which verified solutions best prepare
a model for RL.

\section*{AI Use Statement}
We used generative AI tools to assist with polishing the writing; the authors verified all content and take full responsibility for it. The synthetic data are generated from open source language models. 




\FloatBarrier
\bibliography{main}

\onecolumn
\appendix
\let\AppendixSection\section
\renewcommand{\section}{\FloatBarrier\AppendixSection}
\clearpage
\raggedbottom

\definecolor{AppendixTablePurple}{HTML}{640D5F}
\definecolor{AppendixTableInk}{HTML}{4B0947}
\definecolor{AppendixTableHeader}{HTML}{F1E8F0}
\definecolor{AppendixTableStripe}{HTML}{FAF7F9}
\definecolor{AppendixTableRule}{HTML}{DDD6DC}
\newcommand{\AppendixTableStyle}{%
  \arrayrulecolor{AppendixTablePurple!45!white}%
  \rowcolors{2}{AppendixTableStripe}{white}%
}


\section{Method and Construction Details}
\label{app:method}

\label{sec:method}

The selection pipeline is defined in Section~\ref{sec:topology}.
Numerical table results are rounded to two significant digits; counts,
identifiers and experimental settings remain exact. Differences and ratios
are calculated before rounding.
Within each comparison, the conditions match in candidate eligibility, SFT
trajectory budget, student initialization, training recipe, evaluation
protocol, and checkpoint step. The intended change is the retained solutions or, in a teacher-source
sweep, the generators supplying those solutions. Different testbeds have
different recipes; matching applies within a comparison. The descriptions below
separate candidate generation, verification, representation, and selection.

\paragraph{Generator rosters.}
\label{app:rosters}
In the synthetic environments and for the four-domain Dolci-Think
comparison, open reasoning models write the candidate solutions. For
OpenThoughts3, INTELLECT-3 and Nemotron-Cascade 2,
selection runs directly over the solutions those corpora release. The Dolci-Think pool for
the pre-RL diagnostic (Appendix~\ref{app:dolci-checkpoints}) is written by twelve models, each
answering the same fixed prompts in full. They are Qwen3-32B, Qwen3-14B,
Qwen3-8B, Qwen3-4B-Thinking-2507, DeepSeek-R1-Distill-Qwen-32B,
DeepSeek-R1-Distill-Qwen-14B, DeepSeek-R1-0528-Qwen3-8B,
OpenReasoning-Nemotron-32B, OpenReasoning-Nemotron-14B,
OpenReasoning-Nemotron-7B, AceReason-Nemotron-14B, and AceReason-Nemotron-7B.
In the mathematics-transfer and DAPO teacher-source sweeps, the teacher
order is DeepSeek-R1-0528-Qwen3-8B, Qwen3-8B,
Qwen3-4B-Thinking-2507, OpenReasoning-Nemotron-7B, and Olmo-3-7B-Think, and a
condition with $d\le 5$ teachers uses the first $d$ of them.

The Enigmata teacher-source grid, the twelve-teacher comparison with
Qwen3-14B, and the Qwen3-4B mathematics comparison
(Appendices~\ref{app:construction} and~\ref{app:math-qwen4b-sources}) use
the following ordered twelve-teacher roster:
DeepSeek-R1-0528-Qwen3-8B, Qwen3-8B, Qwen3-4B-Thinking-2507,
OpenReasoning-Nemotron-7B, Olmo-3-7B-Think, Qwen3-14B, Qwen3-32B,
DeepSeek-R1-Distill-Qwen-14B, DeepSeek-R1-Distill-Qwen-32B,
DeepSeek-R1-Distill-Qwen-7B, DeepSeek-R1-Distill-Qwen-1.5B, and
OpenReasoning-Nemotron-32B. In the teacher-count sweeps, the one-, six-, and
twelve-teacher constructions use the first one, six, and twelve entries,
respectively. The separate 14B comparator uses Qwen3-14B alone.

Most comparisons select both conditions from one shared candidate
pool, so both draw on the same generators. The teacher-source comparisons
of Section~\ref{sec:modelcount} and Appendix~\ref{app:construction},
the three Qwen3 points of
Figure~\ref{fig:omega-models}, and the
program-simulation panel of Figure~\ref{fig:olmo-program} instead contrast
corpora from different generators. For the Qwen3 OMEGA
points, Qwen3-4B writes the single-model corpus, and several open reasoning
models write the multi-model corpus by continuing one another's partial
responses. The two corpora of each pair have equal SFT trajectory counts. For program
simulation, the single-model pool holds eight samples per prompt from Qwen3-8B,
and in the multi-model pool several models take turns continuing each response.

\subsection{Problem setup}

Table~\ref{tab:notation} collects the notation.

\begin{table}[!htbp]
\caption{Notation for paired route selection.}
\centering
\small
\AppendixTableStyle
\begin{tabularx}{\linewidth}{p{0.14\linewidth}Y}
\toprule
\rowcolor{AppendixTableHeader}
\bfseries\color{AppendixTableInk}Symbol & \bfseries\color{AppendixTableInk}Meaning \\
\midrule
$x,\ y$ & prompt, candidate solution trajectory (route) \\
$\verifier(x,y)$ & environment verifier, $1$ iff $y$ solves $x$ \\
$\cand^+$ & pool of \emph{verified} candidate routes \\
$\fingerprint(y)$ & topology fingerprint (abstracted transition signature) of route $y$ \\
$d(\cdot,\cdot)$ & distance in fingerprint space \\
$\Ddiv,\ \Dsim$ & matched diverse / similar SFT conditions, with $|\Ddiv|=|\Dsim|$ \\
\bottomrule
\end{tabularx}
\arrayrulecolor{black}

\label{tab:notation}
\end{table}

Let $x$ be a prompt and $\cand_x=\{y_i\}$ a pool of candidate
solutions. A verifier $\verifier(x,y)\in\{0,1\}$ marks whether a solution solves
the prompt, and selection uses only the verified candidates
$\cand_x^+=\{y\in \cand_x:\verifier(x,y)=1\}$. For the three released corpora
the pool is the published solutions
with a complete reasoning span. Verification is applied when constructing the
synthetic and in-house Dolci-Think pools; selecting from a released corpus uses
its existing solutions. Each solution in the pool maps to a
fingerprint $\fingerprint(y)$, and a distance
$d(\fingerprint(y_i),\fingerprint(y_j))$ measures how far apart two routes are.

At a fixed budget $n$, the diverse condition $\Ddiv$ approximately
maximizes the spread of its fingerprints and the similar condition $\Dsim$
approximately minimizes it.
\begin{align}
\Ddiv &\approx \arg\max_{S\subseteq \cand^+, |S|=n}
       \operatorname{Spread}\{\fingerprint(y):y\in S\},\\
\Dsim &\approx \arg\min_{S\subseteq \cand^+, |S|=n}
       \operatorname{Spread}\{\fingerprint(y):y\in S\}.
\end{align}
Both datasets then receive the same SFT recipe and the same RL
recipe, and the reported quantity is the post-RL difference
\[
  \Delta_k =
  \mathbb{E}[\pass{k}\mid \mathrm{SFT}(\Ddiv), \mathrm{RL}]
  -
  \mathbb{E}[\pass{k}\mid \mathrm{SFT}(\Dsim), \mathrm{RL}],
\]
estimated on the same held-out prompts with the same number of
samples. Figures give $\Delta_k$ in points or relative to the similar
condition.

\subsection{Shared selection and training procedure}
\label{app:selection-method}

Every route-selection comparison uses one procedure. Datasets differ only in
the step vocabulary, the feature blocks, the projection, the grouping unit and
the number of clusters (Table~\ref{tab:selection-instances}).

\paragraph{Notation and problem.}
Let $\cand=\{(x_i,y_i,d_i)\}_{i=1}^{N}$ be a candidate pool: prompt $x_i$,
candidate solution $y_i$, and group label $d_i\in\mathcal{G}$. The group is a
domain for the released corpora, Dolci-Think and the single-model pool, and the
shared pool itself when selection runs over the whole pool. The eligible set
$E\subseteq\cand$ holds the candidates a dataset admits: verified candidates
with $\verifier(x_i,y_i)=1$ for the synthetic, OMEGA, Dolci-Think and
single-model pools, and single-turn released solutions (one user and one
assistant turn) with a complete reasoning span for the released corpora. Let
$E_d=\{i\in E: d_i=d\}$. Given a budget $n$, the method returns a diverse set
$\Ddiv\subset E$ with $|\Ddiv|=n$ and a matched low-diversity control $\Dsim$
of the same size. Each set is used for supervised fine-tuning of a fixed
student $\pi_\theta$, followed by reinforcement learning with the same verifier
reward $\verifier$. The measured outcomes are \acc{} and \pass{k} after
reinforcement learning.

\paragraph{Fingerprint.}
A rule-based map $\fingerprint:y\mapsto\fingerprint(y)$ is built in four steps.
\begin{enumerate}[leftmargin=*,itemsep=2pt]
\item \emph{Segmentation and labelling.} The reasoning span of $y$ is split
into steps $\sigma(y)=(s_1,\dots,s_m)$ at blank lines, discourse markers and
sentence boundaries, and a first-match rule assigns each step a type
$\ell(s_t)\in T$ from the dataset's vocabulary. The released-corpus,
Dolci-Think and single-model fingerprints use ten types: setup, computation,
deduction, verification, backtracking, exploration, backward reasoning,
decomposition, commentary and conclusion. RLVE uses 18 note types: correct,
check, branch, parse, answer, infer, manipulate, calculate, enumerate, execute,
insight, decompose, describe, evaluate, assert, track progress, hypothesize and
reason. OMEGA reads 15 annotated step types (setup, conclude, count, matrix,
case, simplify, solve equation, factor, geometry, bound, modular, substitute,
expand, differentiate and integrate), one of 32 strategy labels, and 13 cue
features for checking, backtracking, doubt, switching approach, contradiction
and enumeration.
\item \emph{Graph and tree.} The type sequence $(\ell(s_1),\dots,\ell(s_m))$
induces a directed transition graph $G(y)$ on $T$ with edge weight
$w(a,b)=\#\{t:\ell(s_t)=a,\ \ell(s_{t+1})=b,\ a\neq b\}$, and a reasoning tree
$\tau(y)$ built by a cursor. An ordinary step attaches to the current node by a
sequential edge and becomes current. An exploration step opens a branch. A
verification step attaches as a leaf without moving the cursor. A backtracking
step re-attaches under an earlier node.
\item \emph{Feature vector.} The raw vector concatenates up to five blocks,
\[
f(y)=\big[f_{\mathrm{cont}}(y);\ f_{\mathrm{tree}}(y);\ f_{\mathrm{pat}}(y);\
f_{\mathrm{dense}}(y);\ f_{\mathrm{hash}}(y)\big]\in\mathbb{R}^{D}.
\]
$f_{\mathrm{cont}}$ holds type frequencies, transition rates,
cognitive-behaviour counts, graph topology, step statistics, motifs and tree
shape. $f_{\mathrm{tree}}$ holds per-edge-type transition matrices,
depth-tiered distributions and hashed root-to-leaf path signatures.
$f_{\mathrm{pat}}$ holds binary strategy-pattern indicators.
$f_{\mathrm{dense}}$ holds conversation-level lengths, entropies and marker
positions. $f_{\mathrm{hash}}$ is a signed feature-hashing bag of route labels
(type $n$-grams, tree node and edge labels, root-to-leaf paths and attempt
sequences), where SHA-256 of each label gives its index and sign, and the
vector is scaled by one over the square root of the number of labels.
Table~\ref{tab:selection-instances} lists the blocks each dataset uses.
\item \emph{Standardization and projection.} With pool mean $\mu_j$ and
standard deviation $\sigma_j$ of coordinate $j$, computed on the eligible pool
before either condition is selected,
\[
z_j(y)=\operatorname{clip}\!\Big(\tfrac{f_j(y)-\mu_j}{\sigma_j},-5,5\Big),\qquad
\fingerprint(y)=\frac{\hat z(y)\,P}{\lVert \hat z(y)\,P\rVert_2},\quad
\hat z=\frac{z}{\lVert z\rVert_2},
\]
where $P\in\mathbb{R}^{D\times r}$ has independent $\mathcal{N}(0,1/r)$ entries
drawn once from a fixed seed. Datasets without a projection use
$\fingerprint(y)=\hat z(y)$, and a zero vector stays at zero. A constant
feature uses $\sigma_j=1$. RLVE standardizes its continuous and topology blocks
and appends the 64 binary pattern indicators unscaled. Distances are Euclidean,
$\delta(i,j)=\lVert\fingerprint_i-\fingerprint_j\rVert_2$.
\end{enumerate}

\paragraph{Selection.}
Write $\fingerprint_i=\fingerprint(y_i)$.
\begin{enumerate}[leftmargin=*,itemsep=2pt]
\item \emph{Group budgets.} The quotas $n_d$ are the most even split of $n$
subject to $n_d\le|E_d|$ and $\sum_d n_d=n$. Every group receives
$\min(|E_d|,q)$ for a common level $q$, and any remainder goes one row at a
time to the groups with the most remaining capacity. With a single group,
$n_d=n$.
\item \emph{Clustering.} Within group $d$, a cluster count
$K_d=\max\!\big(1,\min(\lfloor K\,n_d/n\rceil,\ |E_d|,\ n_d)\big)$ is derived
from a shared total $K$, and mini-batch $k$-means with a fixed seed
approximately minimizes
\[
\sum_{i\in E_d}\lVert\fingerprint_i-m_{c(i)}\rVert_2^2
\]
over assignments $c:E_d\to\{1,\dots,K_d\}$ and centres $m_c$, giving clusters
$C_c=\{i:c(i)=c\}$.
\item \emph{Cluster quotas.} Each nonempty cluster receives
\[
b_c=\max\!\Big(1,\ \Big\lfloor \tfrac{|C_c|}{|E_d|}\,n_d\Big\rfloor\Big),
\]
capped at $|C_c|$, and the remaining $n_d-\sum_c b_c$ rows are assigned by
largest fractional remainder so that $\sum_c b_c=n_d$ exactly.
\item \emph{Within-cluster spread.} For a cluster with $b_c<|C_c|$, greedy
farthest-point sampling picks
\[
S_c^{(0)}=\Big\{\arg\min_{i\in C_c}\lVert\fingerprint_i-\bar{\fingerprint}_c\rVert_2\Big\},\qquad
S_c^{(t+1)}=S_c^{(t)}\cup\Big\{\arg\max_{i\in C_c\setminus S_c^{(t)}}\
\min_{j\in S_c^{(t)}}\delta(i,j)\Big\},
\]
until $|S_c|=b_c$, with $\bar{\fingerprint}_c$ the cluster mean. This is the
greedy $k$-center construction \citep{coreset2018}, a spread heuristic that
approximates the set objective above, and squared and plain Euclidean distances
give the same picks. Clusters with $b_c=|C_c|$ are taken whole.
\item \emph{Output and control.} $\Ddiv=\bigcup_d\bigcup_c S_c$, so
$|\Ddiv|=n$. The control $\Dsim$ keeps, per group, the $n_d$ rows with the
smallest $\lVert\fingerprint_i-\mu_d\rVert_2$, where $\mu_d$ is the group mean
of $\fingerprint$. The $k$-means seed is the only random input. Steps 1, 3 and 4
and the control are deterministic given the clustering. The two selections
need not be disjoint.
\end{enumerate}

\paragraph{Training.}
Supervised fine-tuning minimizes the token-level negative log-likelihood of
the response,
\[
\mathcal{L}_{\mathrm{SFT}}(\theta)=-\sum_{(x,y)\in\mathcal{D}}\ \sum_{t}\log\pi_\theta(y_t\mid x,y_{<t}),
\]
over the kept rows. Reinforcement learning uses group relative policy
optimization from the fine-tuned checkpoint $\pi_{\mathrm{ref}}$. For each
prompt $x$ a group of $G$ responses is sampled, each receives the binary reward
$r_g=\verifier(x,y_g)$, the advantage is group-normalized,
$\hat A_g=(r_g-\bar r)/\mathrm{std}(r)$, and the actor maximizes the clipped
importance-weighted surrogate with a per-token Kullback-Leibler penalty toward
$\pi_{\mathrm{ref}}$, using the low-variance estimator,
\[
\mathcal{J}(\theta)=\mathbb{E}\Big[\tfrac{1}{G}\sum_{g=1}^{G}\tfrac{1}{|y_g|}\sum_{t}\min\!\big(\rho_{g,t}\hat A_g,\ \operatorname{clip}(\rho_{g,t},1-\epsilon,1+\epsilon)\hat A_g\big)\Big]-\beta\,\mathbb{D}_{\mathrm{KL}}\!\big[\pi_\theta\,\Vert\,\pi_{\mathrm{ref}}\big],
\]
with $\rho_{g,t}=\pi_\theta(y_{g,t}\mid\cdot)/\pi_{\theta_{\mathrm{old}}}(y_{g,t}\mid\cdot)$
and no entropy term. Prompts whose group rewards are all equal have zero
advantage and contribute no reward-driven update. Both conditions of a
comparison receive the same SFT and RL recipe (Table~\ref{tab:config}).

\paragraph{Evaluation.}
For problem $i$ and $k$ samples $y_{i1},\dots,y_{ik}$ with
$r_{ij}=\verifier(x_i,y_{ij})$,
\[
\text{\acc}=\frac{1}{M}\sum_{i=1}^{M}\frac{1}{k}\sum_{j=1}^{k}r_{ij},\qquad
\text{\pass{k}}=\frac{1}{M}\sum_{i=1}^{M}\mathbf{1}\Big[\max_j r_{ij}=1\Big],
\]
reported per benchmark and as the unweighted mean over a comparison's
benchmarks. When more than $k$ samples are drawn, Appendix~\ref{app:metrics}
gives the estimator used for \pass{k}.

\paragraph{Proposed account.}
If the policy solves problem $x$ with per-sample probability $p(x)$, then
$\text{\pass{k}}(x)=1-(1-p(x))^k$. Route-diverse fine-tuning is proposed to
raise the number of problems with $p(x)>0$, which raises \pass{k} directly,
and group-relative reinforcement learning updates only on prompts with
$0<\bar r<1$, so problems newly within sampling reach become the ones
reinforced. The link from repertoire to $p(x)$ is the part of this account we
do not measure. Appendix~\ref{app:model} develops it as an illustrative model.

\begin{table}[!htbp]
\caption{How each dataset instantiates the shared procedure.}
\label{tab:selection-instances}
\centering
\small
\setlength{\tabcolsep}{4pt}
\AppendixTableStyle
\begin{tabularx}{\linewidth}{>{\raggedright\arraybackslash}p{0.15\linewidth} Y Y >{\raggedright\arraybackslash}p{0.13\linewidth} >{\raggedright\arraybackslash}p{0.14\linewidth}}
\toprule
\rowcolor{AppendixTableHeader}
Dataset & Groups & Feature blocks (raw dimension) & Projection & Clusters \\
\midrule
RLVE & shared RLVE pool (64 environments for OLMo3-7B) & continuous 159, topology 1{,}150, pattern 64 (1{,}373) & Gaussian & proportional budgets \\
OMEGA & accepted OMEGA pool & 323 strategy-step features (Appendix~\ref{app:omega-construction}) & none & proportional budgets \\
Dolci-Think & four domains, 25{,}000 rows each & continuous 137, tree 430, pattern 64 (631) & 96, seed 42 & 40{,}000 \\
Single-model pool & domains, at most eight solutions per prompt & continuous 137, tree 430, pattern 64 (631) & 96, seed 42 & within each domain \\
Released corpora & domains & continuous 137, tree 430, pattern 64, dense 83, hash 1{,}024 (1{,}738) & 96, seed 42 & 40{,}000 \\
Sokoban & shared Sokoban pool & route signatures of the move sequence & none & proportional budgets \\
\bottomrule
\end{tabularx}
\end{table}

\subsection{Evaluation metrics and comparison units}
\label{app:metrics}

Let $c_i$ be the number of correct responses among $N$ sampled responses to
question $i$. Mean sampled accuracy is the mean of $c_i/N$ over questions.
When estimating a smaller sampling budget $k\leq N$ from these same responses,
we use the finite-sample coverage estimator
\[
  \widehat{\pass{k}}=
  \frac{1}{Q}\sum_{i=1}^{Q}
  \left[1-\frac{\binom{N-c_i}{k}}{\binom{N}{k}}\right],
\]
where the numerator is zero if $N-c_i<k$. At $k=1$ this equals mean sampled
accuracy on the same question set; at $k=N$ it is the fraction of questions
solved at least once. Qwen3 RLVE coverage uses the reference-answer subset,
whereas its mean sampled accuracy uses the full scored question set.
Every \pass{1} in the paper, including the Qwen3 RLVE and program-simulation
points, is this mean over sampled responses.
IFEval and IFBench use strict prompt-level instruction-following accuracy.

Question sets, number of draws, decoding settings, output caps, and scoring
rules are the same for both conditions of a comparison. Every question in
the specified evaluation set contributes to its denominator. Different budgets
computed from one set of responses are correlated measurements. An absolute
gap is $100(s_D-s_S)$ percentage points for scores on $[0,1]$; a relative gain
is $100(s_D/s_S-1)\%$. Benchmark means weight benchmarks equally unless the
result is explicitly labeled pooled, in which case questions are weighted
equally. Relative gains of benchmark means are defined in
Appendix~\ref{app:realdata-relative-benchmarks}.

\paragraph{Checkpoints and uncertainty.}
Every post-RL comparison evaluates both conditions at the same RL checkpoint
step. The paired pre-RL reward diagnostics use matched end-of-SFT checkpoints.
Unless an experiment explicitly reports
multiple seeds, each condition uses one training run. Question-level paired
tests and ranges across evaluation settings do not estimate variation across
training seeds. A positive post-RL difference establishes an endpoint advantage;
it does not by itself establish a larger improvement during RL.

\paragraph{Reward-signal diagnostics.}
For a group of $G$ binary rewards with $c$ successes, the outcome is all-fail
if $c=0$, all-correct if $c=G$, and mixed if $0<c<G$. Informative share is the
fraction of prompts in the mixed category. The Dolci-Think diagnostic uses
64 shared mathematics prompts, eight samples per prompt, and temperature 1.0.
The pre-RL RLVE diagnostic uses 8 or 32 samples with response caps of 16,384
or 32,768 tokens, respectively. Both caps are shared within each paired
comparison, so the change between those two diagnostic budgets also changes
the response cap. In Figure~\ref{fig:rlve-olmo-performance}, the post-RL
Seen/Unseen coverage curves use the same budget-dependent caps; the All
coverage curve and solved-set overlap use a 32,768-token cap.

\subsection{Measuring topology spread}
\label{app:spread}

Diversity in curated data is often claimed without being measured
\citep{zhao2024measure}, although general measures exist for generations and
datasets, such as the Vendi score \citep{friedman2023vendi}, self-similarity and
distinct-$n$ measures \citep{zhu2018texygen,tevet2021evaluating}, and dataset
diversity coefficients \citep{miranda2024diversity}. We measure the quantity the
two conditions are built to differ in, which is how far apart the verified
solutions to the same prompt are.

Every statistic is computed among the selected solutions of one
prompt and then averaged over all prompts with at least two selected solutions.
For these descriptive statistics, features are standardized within each
selected pool before distances are taken; this is separate from the shared
candidate-pool normalization used for selection.
\emph{Response pairwise distance} is the mean Euclidean distance between the
159 continuous features of two solutions to the same prompt. \emph{Topology
pairwise distance} is the same mean taken over the 1{,}150 topology features of
the RLVE fingerprint, and \emph{topology-vector variance} is the variance of
those topology features about their mean, averaged over dimensions. Two pattern
statistics use the 64 pattern indicators and record how many reasoning patterns
the solutions use and how evenly. \emph{Pattern
entropy} is the Shannon entropy, in bits, of the pattern occurrences pooled over
the prompt's solutions, and the \emph{active-pattern count} is the number of
patterns that occur in at least 2\% of them.

\paragraph{Controls.}
The following controls separate route spread from other differences between
the conditions. Both conditions of each topology-selected pair come from one
pool $\cand^+$. In the synthetic environments and the four-domain
Dolci-Think comparison, both pass the same verifier or answer checker, and for
the three released corpora both use the released solutions directly. Each paired selection has the
same SFT trajectory budget in both conditions. The RLVE fingerprint reads its
events from the wording of each trace, while the OMEGA fingerprint is built
mostly from annotated steps and strategy labels, and the diverse selection leads
there as well (Figure~\ref{fig:omega-models}).

Table~\ref{tab:calib} applies these statistics to the two
50{,}000-row RLVE selections of the Qwen3-4B pair in
Figure~\ref{fig:rlve-qwen-budget}, over the 5{,}910 diverse and 10{,}482 similar
prompts that keep at least two selected solutions. These are descriptive
within-prompt statistics of each selected dataset: the eligibility rule is
the same, and its qualifying prompt set can differ between selections.
The selections have nearly equal pattern entropy and active-pattern count,
and separate on every distance or variance statistic. This calibration
measures the fingerprint's geometry; it does not count semantically distinct
algorithms.

\begin{table}[!htbp]
\caption{Spread statistics of the 50{,}000-row diverse and
similar RLVE selections (the Qwen3-4B pair at 50{,}000 rows in
Figure~\ref{fig:rlve-qwen-budget}). Per prompt, the two selections use
about the same number of reasoning patterns, spread about as evenly, and the
distance between solutions is about 2.3 times larger in the diverse selection.}
\label{tab:calib}
\centering
\small
\setlength{\tabcolsep}{6pt}
\renewcommand{\arraystretch}{1.25}
\AppendixTableStyle
\begin{tabularx}{0.82\linewidth}{>{\leavevmode\bfseries\color{AppendixTableInk}}Y r r r}
\rowcolor{AppendixTablePurple}
{\color{white}Pool statistic} & {\color{white}Diverse pool} &
{\color{white}Similar pool} & {\color{white}Ratio}\\
Response pairwise distance & 11 & 5.0 & 2.3\\
Topology pairwise distance & 23 & 9.8 & 2.3\\
Topology-vector variance & 0.14 & 0.068 & 2.0\\
Pattern entropy & 4.9 & 4.9 & 1.0\\
Active-pattern count & 32 & 31 & 1.0\\
\bottomrule
\end{tabularx}
\arrayrulecolor{black}

\end{table}

\subsection{RLVE data construction}

\paragraph{Topology fingerprint and selection.}
Each RLVE trace is summarized by a lexical-topological fingerprint of
its reasoning span. The fingerprint has 1{,}373 dimensions, made of 159
continuous features, 1{,}150 topology features, and 64 pattern indicators.
The continuous features describe note-type frequencies, selected transitions,
step lengths, verification and revision, subgoals, and temporal position.
The topology block contains 972 edge-type-specific note transitions (three
$18\times18$ matrices), 54 parent-conditioned edge-type frequencies,
72 depth-conditional summaries over three tiers, 40 hashed root-to-leaf path
features, and 12 tree-shape statistics. The three edge types are sequential
continuation, backtracking, and exploration. Transition and edge-type
frequencies are row-normalized; path-feature counts are normalized by leaf
count. Pattern indicators
cover direct reasoning, verification, exploration, backtracking, decomposition,
task-specific procedures, and failure patterns such as circular reasoning or
abandoned approaches. The continuous and topology blocks are standardized
using shared candidate-pool statistics and clipped to $[-5,5]$, then
concatenated with the 64 binary pattern indicators. These are text-derived abstractions of
the visible reasoning span, not traces of a model's internal computation.
Diverse selection projects and
clusters the candidate pool and runs farthest-point selection inside each
cluster, with budgets proportional to cluster size. Similar selection
keeps the rows nearest the centroid of the pool. At
both Qwen3 model sizes in Figure~\ref{fig:rlve-qwen-budget}, the Diverse and
Similar conditions are selected from the same shared pool with this
procedure, at 50{,}000 and at 200{,}000 rows.

For OLMo3-7B, the shared candidate pool is restricted to 64 RLVE environments,
of which 63 occur in the held-out evaluation. Both conditions retain
50{,}000 SFT rows in total, selected from this 64-environment pool with the
lexical-topological fingerprints and the selection procedure above, with matched
budgets and training.
The Seen/Unseen evaluation split refers to this SFT environment
set, not to the environments available during RL.

\subsection{OMEGA data construction}
\label{app:omega-construction}

\paragraph{Multi-model and single-model candidate generation.}
Single-model trajectories are written by Qwen3-4B alone. Multi-model trajectories are produced by a roster of open
reasoning models from several families, including Qwen, DeepSeek, and Nemotron. The roster passes
each response from model to model in segments of up to 1,024 tokens, and each
model resumes the previous model's assistant turn token for token, with no new
user prompt. The roster includes Qwen3-14B, Nemotron-Cascade-14B-Thinking,
OpenMath-Nemotron-14B, AceReason-Nemotron-14B, Nemotron-Nano-9B-v2,
Phi-4-reasoning-plus, Olmo-3-7B-Think and DeepSeek-R1-Distill-Qwen-7B.
An English instruction to the roster and a filter that drops
any trajectory containing Chinese characters keep the generations in English.

\paragraph{Verification and SFT data construction.}
The OMEGA math verifier checks each generated candidate's answer, and only
accepted candidates enter the verified pool. Selection reads each accepted
trace's recorded strategy-step sequence and strategy label.
The two reported OMEGA comparisons use different corpora. The three Qwen3 points of
Figure~\ref{fig:omega-models} contrast a multi-model corpus with a single-model corpus, and each
pair uses the same SFT trajectory budget. All three are read on a fixed
300-prompt held-out subset. The OLMo3-7B point uses the fingerprint selection described next. It
selects 50{,}000 rows per condition from one shared verified pool that holds candidates from both
generator families, and it is read on a deterministic 500-prompt held-out subset.
Both evaluation subsets come from a 592-problem pool split from RL by problem
ID within each setting. The same settings occur in RL and evaluation, but
their problem IDs are disjoint.

\paragraph{Topology selection.}
For OMEGA we compute the topology fingerprint mostly from the recorded strategy-step sequence
and strategy label of each accepted solution. It includes step unigrams, row-normalized transition
probabilities, features for the first and last thirds of the trace, scale-free scalar features, a
strategy-label one-hot vector, and 13 lexical cue features. The
323-dimensional vector is constructed as follows:
\begin{itemize}
\item 15 step frequencies, normalized by sequence length, and $15\times15=225$
next-step probabilities, normalized within each source-step row. A row with no
outgoing transition is zero.
\item 30 position features: a 15-type frequency distribution for the first
third and another for the last third of the step sequence.
\item Eight path summaries: the number of distinct step types divided by
sequence length and by 15; transition entropy divided by $\log(225)$;
the longest repeated-step run divided by sequence length; the number of
distinct transitions divided by the number of transitions; the fraction of
self-transitions; a strategy-switch indicator; and the fraction of observed
transition types that recur. Empty denominators contribute zero.
\item A 32-dimensional one-hot strategy label, followed by six cue-composition
fractions, six late-cue fractions, and one log-density feature. The cue families
are verification, backtracking, uncertainty, an alternative approach,
contradiction, and enumeration. Composition divides a family's hits by all cue
hits; its late fraction is the share in the second half of the response's
character positions. The density is
$\log(1+1000C/\max(1,W))$, with $C$ cue hits and $W$ words. A family with no
hits has zero late fraction.
\end{itemize}
We retain accepted solutions with a boxed answer and usable step annotations,
and remove exact duplicate response texts. A missing or unrecognized strategy
label contributes an all-zero strategy-label block. Features are standardized
over this eligible pool and L2-normalized before Euclidean distances are
computed. From that pool, diverse and similar selection keep
equal numbers of trajectories using clustered farthest-point and nearest-centroid
selection, respectively.

The strategy vocabulary comprises modular arithmetic, Euclidean GCD,
coordinate geometry, case split, row reduction, substitution,
inclusion--exclusion, symbolic simplification, brute-force enumeration,
prime factorization, equation solving, de Moivre's formula, generating
functions, monotonicity, complex numbers, dynamic programming, bounding,
direct algebra, graph search, block decomposition, linear dependence,
roots of unity, symmetry, rank factorization, complement counting,
invariants, algorithmic generalization, outer-product decomposition,
inversion, synthetic geometry, power of a point, and other.

\subsection{Sokoban data construction}
\label{app:sokoban-construction}

\paragraph{Boards and held-out evaluation.}
The held-out boards come from the 10{,}000-record evaluation split of a
Sokoban corpus in which every record was replay-verified during generation and
again in an independent pass, and no board-answer pair appears in more than one
split. The reported comparison is read on a fixed 500-board subset of the evaluation split, with 193 easy, 176
medium, 96 hard, and 35 expert boards, at \pass{1}, \pass{4}, and \pass{64}.

\paragraph{Route fingerprint.}
Sokoban fingerprints are computed from the visible reasoning trace and the moves it describes. One group of signatures records how a trace travels over the board, for example whether it walks straight to a target, backtracks, loops, keeps returning to one central cell, lists cells along a row or column, follows a corridor, sweeps a room, or spreads outward from a start. A second group records how it handles the boxes, for example whether it starts from the goals, starts from a box and pushes forward, pairs boxes with goals, tests the line a box can be pushed along, or checks for positions where a box would be stuck. Many traces spend much of their length rebuilding the board, so a third group records how they do it, for example by repeating the same coordinates, re-reading walls or landmarks, locating the objects, listing the board state briefly, or correcting an earlier reading.
Graph-shape and transition signatures computed from the path of board
coordinates that each trace visits complete the fingerprint.

\paragraph{Selection.}
Sokoban uses the shared procedure of Appendix~\ref{app:selection-method}. The
diverse condition clusters the fingerprints, gives each cluster a share of the
budget in proportion to its size, and runs farthest-point selection inside
each cluster. The similar condition keeps the traces nearest the centroid of
the pool. Both conditions select 86{,}792 rows from one shared pool of verified traces.
Both conditions then run the same 75-step GRPO with a solved-only reward on one
set of 10{,}591 boards drawn at random from a separately generated pool of
120{,}000 boards, most of them easy or medium (Table~\ref{tab:config}). This is the pair
reported in Figure~\ref{fig:olmo-sokoban}.

\subsection{Program-simulation data construction}

\paragraph{Task and verification.}
Each program-simulation prompt gives a short program in a rewrite system adapted from the A::B environment of RLVE \citep{zeng2026rlve} and asks for the state the program ends in. An exact checker compares the answer with the reference final state, and only trajectories it accepts enter a corpus.

\paragraph{The two corpora.}
Figure~\ref{fig:olmo-program} compares two SFT corpora of 10{,}000 verified trajectories each, drawn with a fixed seed from their respective pools. The figure labels the multi-model corpus Diverse and the single-model corpus Similar, and Appendix~\ref{app:rosters} describes how each is generated.

Multi-model generation deterministically shuffles its active roster for each
prompt and candidate, then cycles through that order in segments of at most
1,024 tokens. Each teacher resumes the accumulated assistant response using
its chat template. The generator roster includes Qwen3.5-2B, -4B, -9B and
-27B, Qwen3.6-27B, AceReason-Nemotron-7B and -1.1-7B, OpenThinker2-7B and
-32B, and OpenThinker3-1.5B and -7B. Generation uses subsets of this roster;
an individual trajectory mixes up to four teachers. Sampling uses temperature
0.8 and top-$p=0.95$, with a total 16,384-token cap, and stops when a complete
answer tag is produced. The single-model corpus uses Qwen3-8B to generate
eight candidates per prompt. The checker is shared by both generator pools.

\paragraph{Training and evaluation.}
OLMo3-7B receives 300 full-parameter SFT updates at batch size 32 on each corpus. Both students then run the same 75-step GRPO on one shared set of RL prompts (Table~\ref{tab:config}). Evaluation uses 500 held-out prompts balanced across seven difficulty levels, from tiny to extreme, with 64 samples per prompt. The one-, four- and 64-sample points are \pass{k} over the 64 samples drawn at temperature 0.8, so \pass{1} is the mean sampled accuracy.

\section{Supporting Construction Sweeps}
\label{app:construction}

The teacher-source sweeps vary the number of generators at a fixed
SFT trajectory budget, and a separate sweep varies the size of the task pool.
Table~\ref{tab:construction-config} gives their protocols. Section~\ref{sec:modelcount}
shows the main results in Figures~\ref{fig:mot-route-and-source}
and~\ref{fig:mot-route-source}, and this appendix gives the task-pool figure
and the complete sweeps.
Within each teacher-source comparison, the student, prompt pool, retained
SFT trajectory budget, training recipes, evaluation protocol, and checkpoint
step are matched. The
teacher count $d$ changes which generators supply the solutions at that
budget. For the one- through five-teacher source, transfer and DAPO sweeps, the
condition with $d$ teachers uses the first $d$ entries of the fixed roster in
Appendix~\ref{app:rosters}. Thus source identity changes along with source
count; teacher count is a proxy for route diversity.

\begin{table}[!htbp]
\centering
\caption{Protocols for the supporting construction sweeps. Teacher
count is the number of generators that supply solutions across the corpus.}
\label{tab:construction-config}
\small
\AppendixTableStyle
\begin{tabularx}{\linewidth}{>{\raggedright\arraybackslash}p{.17\linewidth} Y Y}
\toprule
\rowcolor{AppendixTableHeader}
Sweep & Construction & Evaluation \\
\midrule
OMEGA source sweep & Qwen3-1.7B; one through five teachers over RLVE pools of 16 and 399 environments; RL on OMEGA's training set. & RL step 350; 265 OMEGA compositional problems from its out-of-distribution set; 64 samples per problem (Figures~\ref{fig:mot-ood-gain} and~\ref{fig:app-mot-ood-full}). \\
Task pool & Qwen3-1.7B, ten reasoning-gym task-pool sizes from 2 to 92 tasks. & Pre-RL Enigmata evaluation, 93 held-out problems and 256 samples per problem. Three seeds per pool size. \\
Transfer sweeps & Qwen3-1.7B; one through five teachers over RLVE pools of 16 and 399 environments; RL on reasoning-gym tasks. & RL step 350; 265 OMEGA compositional problems and four mathematics benchmarks; 64 samples per problem, temperature 1.0 and an 8,192-token cap (Figures~\ref{fig:app-response-diversity-omega-full} and~\ref{fig:app-response-diversity-math-full}). \\
Qwen3-4B mathematics & One and twelve teachers over RLVE pools of 16 and 399 environments; RL on DAPO-Math-17k. & Four mathematics benchmarks, RL step 200, 64 samples per problem, 8,192-token cap (Appendix~\ref{app:math-qwen4b-sources}). \\
Enigmata teacher grid & Qwen3-1.7B and Qwen3-4B; one, six, or twelve teachers over RLVE pools of 16 and 399 environments; RL on Enigmata. & Held-out Enigmata, RL step 750 (1.7B) and 500 (4B). The 4B evaluation has 486 problems and 64 samples per problem (Figure~\ref{fig:mot-route}). \\
Enigmata decomposition & Qwen3-1.7B; twelve retained solutions per prompt, from one or twelve teachers; RL on Enigmata. & Sampled \pass{k} at $k\in\{1,8,32,64\}$ on 125 in-domain problems; direct RL provides a third baseline (Figure~\ref{fig:mot-enigmata-decomp}). \\
14B-teacher comparison & Qwen3-1.7B student; verified solutions from one Qwen3-14B teacher or twelve teachers on the same prompts at a matched SFT budget; RL on Enigmata's twelve-task training set. & 125 in-domain and 361 out-of-domain Enigmata problems; 64 samples per problem, temperature 1.0 and an 8,192-token cap (Figure~\ref{fig:mot-teacher-route}). \\
Mathematics RL companion & The transfer sweep's Qwen3-1.7B SFT conditions; RL on DAPO-Math-17k. & Four mathematics benchmarks, RL step 50, 64 samples per problem, 8,192-token cap (Appendix~\ref{app:dapo}). \\
\bottomrule
\end{tabularx}
\end{table}

In the one- through five-teacher sweeps, each difference subtracts that
comparison's $d{=}1$ score from its $d{=}2,\ldots,5$ score. In the sweep with
RL on reasoning-gym tasks, the OMEGA compositional and mathematics-transfer
results use the same step-350 checkpoints, and the pool-16 and
pool-399 one-teacher \pass{64} baselines on OMEGA compositional are $34.34\%$
and $33.58\%$, respectively. The OMEGA source sweep, with RL on OMEGA's training set,
is a separate comparison with its own one-teacher baselines, $37.36\%$ in pool 16
and $33.96\%$ in pool 399.
Values are absolute pass-rate differences unless noted.
The teacher-source results use one training run per condition. Budgets
computed from the same generated samples are correlated readouts; the
reported win counts count metric cells.

\begin{figure}[!htbp]
\centering
\includegraphics[width=4.4in]{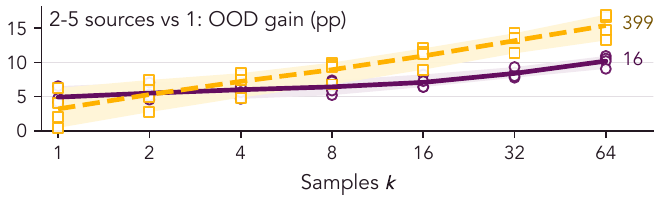}
\caption{Complete OMEGA source sweep behind Figure~\ref{fig:mot-ood-gain}. Marks
are post-RL gains in OMEGA compositional coverage, in points, of two- through
five-teacher conditions over the one-teacher condition, for SFT pools of 16 and
399 environments. RL trains on OMEGA's training set. All 56 comparisons are
positive. Lines are means and bands observed ranges.}
\label{fig:app-mot-ood-full}
\end{figure}

\paragraph{OMEGA source sweep.}
Multi-teacher mixtures outperform the single-teacher corpus at the same
trajectory budget when RL trains on OMEGA's training set. After RL, at RL step
350, five-teacher corpora exceed one-teacher corpora on OMEGA compositional
coverage at every sampling budget in both environment pools
(Figure~\ref{fig:mot-ood-gain}). At \pass{64} the five-teacher corpus reaches
$46.4\%$ against $37.4\%$ in the 16-environment pool and $47.2\%$ against
$34.0\%$ in the 399-environment pool. The complete sweep adds two, three and
four teachers, and all 56 comparisons of a multi-teacher condition with the
one-teacher condition are positive (Figure~\ref{fig:app-mot-ood-full}). The
mean gain over one teacher grows with the sampling budget, from 4.9 points at
\pass{1} to 10.2 points at \pass{64} in the 16-environment pool and from 3.2 to
15.4 points in the 399-environment pool. \pass{64} peaks at two teachers in
pool 16 ($48.30\%$) and at four teachers in pool 399 ($50.94\%$).

\begin{figure}[htbp]
\centering
\includegraphics[width=2.75in]{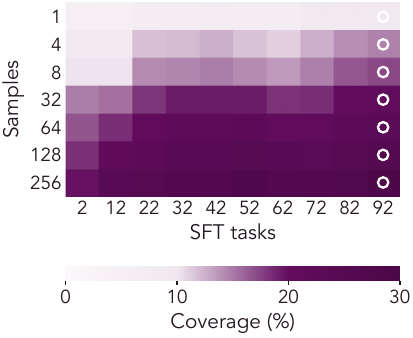}
\caption{Held-out Enigmata coverage in percent (shading) for Qwen3-1.7B
after SFT on pools of 2 to 92 reasoning-gym tasks.
Coverage generally rises, with diminishing gains and local reversals.
White rings mark each sampling budget's best pool. Three-seed means, before RL.}
\label{fig:mot-taskcount-coverage}
\end{figure}

\paragraph{Task-pool size.}
Before RL, held-out Enigmata coverage of a Qwen3-1.7B student generally rises as the SFT task pool grows from 2 to 92 tasks, with diminishing gains and local reversals (Figure~\ref{fig:mot-taskcount-coverage}). Much of the gain arrives by 22 to 42 tasks. \pass{64} rises by 6.5 points between the smallest and the largest pool. The 62-task pool trails the 52-task pool at every reported sampling budget; at \pass{64}, coverage falls from $22.10\%$ to $20.70\%$. The three-seed ranges at \pass{64} are $16.61$--$16.79\%$ for two tasks and $22.63$--$23.86\%$ for 92 tasks; these are observed ranges, not confidence intervals.

\paragraph{Twelve teachers against one.}
On Enigmata, a Qwen3-1.7B student trained on twelve solutions per prompt from twelve generators reaches $35.2\%$ \pass{64} on the 125 in-domain problems after RL. The same student reaches $28.0\%$ when all twelve solutions come from one generator, and $12.8\%$ with RL and no SFT (Figure~\ref{fig:mot-enigmata-decomp}). The figure reports sampled coverage at $k\in\{1,8,32,64\}$; the horizontal axis is the number of sampled completions per problem.

The Enigmata teacher-source comparisons use verified solutions to the same
RLVE prompts within each pair. The one-, six- and twelve-teacher conditions
use the first one, six and twelve generators in the ordered roster of
Appendix~\ref{app:rosters}, with the retained trajectory budget held fixed.
The decomposition in Figure~\ref{fig:mot-enigmata-decomp} additionally fixes
the retained count at twelve solutions per prompt in both SFT conditions.

\Needspace{12\baselineskip}
\paragraph{Comparison with a single Qwen3-14B teacher.}
Figure~\ref{fig:mot-teacher-route} compares two complete SFT--RL recipes
for a Qwen3-1.7B student. Both recipes use verified solutions to the same
RLVE prompts, with the same retained SFT trajectory budget and matched
training checkpoints. One recipe draws its solutions from Qwen3-14B alone;
the other draws from the twelve-teacher roster in Appendix~\ref{app:rosters}.
Both recipes train RL on Enigmata's
twelve-task training set. Evaluation uses 125 problems from the training
task families (ID) and 361 from held-out
task families (OOD), with the decoding settings in
Table~\ref{tab:construction-config}. The comparison measures the outcome
of the complete recipes, rather than isolating teacher size or source count.

\begin{figure}[!htbp]
\centering
\includegraphics[width=4.4in]{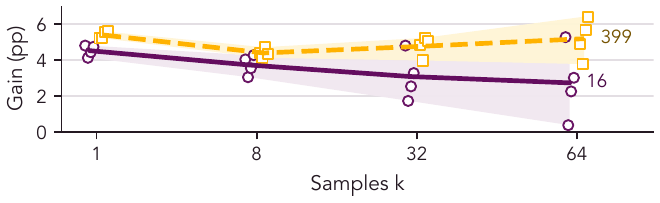}
\caption{Full OMEGA compositional profile behind
Figure~\ref{fig:response-diversity-omega-transfer}, after RL (step 350), with $d$ denoting teacher count.
Open marks give every
$d{=}2,\ldots,5$ coverage difference from $d{=}1$ in points at each budget $k$, for SFT
pools 16 (circles) and 399 (squares). RL trains on reasoning-gym
tasks, and evaluation uses OMEGA's out-of-distribution compositional problems,
which lie outside both training stages. All 32 comparisons are positive. Lines are means and bands
observed ranges.}
\label{fig:app-response-diversity-omega-full}
\end{figure}

\paragraph{Transfer to out-of-distribution problems and mathematics.}
The teacher-source advantage also holds on OMEGA's out-of-distribution problems, which combine skills beyond OMEGA's training distribution. With Qwen3-1.7B trained by SFT on corpora from one to five teachers at equal total trajectory counts, then given the same RL on reasoning-gym tasks, so that neither stage trains on OMEGA, every multi-teacher condition exceeds the single-teacher condition after RL, at RL step 350, on OMEGA's out-of-distribution compositional evaluation set in both SFT environment pools and at every reported sampling budget (Figures~\ref{fig:response-diversity-omega-transfer} and~\ref{fig:app-response-diversity-omega-full}).

\begin{figure}[!htbp]
\centering
\includegraphics[width=4.4in]{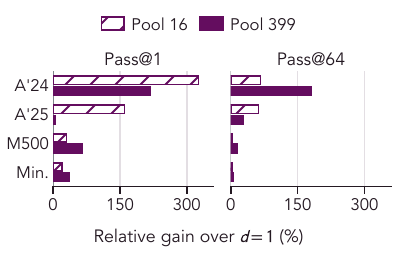}
\caption{Mathematics transfer for Qwen3-1.7B after RL on reasoning-gym tasks, at saved
step 350. Five teacher sources are compared with one at equal total SFT
trajectory counts, with one training run per condition. Evaluation uses 64
draws per problem, temperature 1.0, and an 8,192-token cap. Bars show relative gain,
$100(\mathrm{score}_{d=5}-\mathrm{score}_{d=1})/\mathrm{score}_{d=1}$ in percent.
\pass{1} is left and \pass{64} right. Hatched and solid bars denote pools 16 and
399. A'24, A'25, M500, and Min. denote AIME 2024, AIME 2025, MATH-500, and Minerva.
These four benchmarks are not part of either training set.
Figure~\ref{fig:app-response-diversity-math-full} gives every teacher-count level.}
\label{fig:response-diversity-math-transfer}
\end{figure}

\begin{figure}[!htbp]
\centering
\begin{subfigure}[t]{\linewidth}
\centering
\includegraphics[width=4.4in]{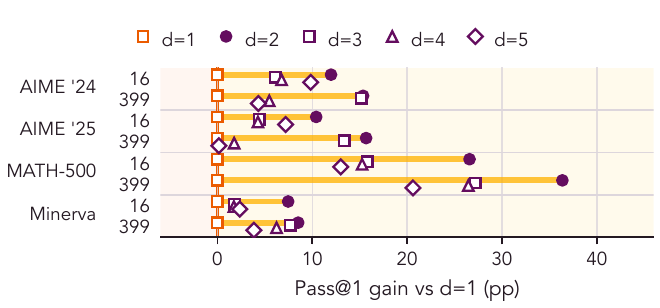}
\caption{Complete mathematics ladder at \pass{1}.}
\label{fig:app-math-xfer-full-pass1}
\end{subfigure}
\par\medskip
\begin{subfigure}[t]{\linewidth}
\centering
\includegraphics[width=4.4in]{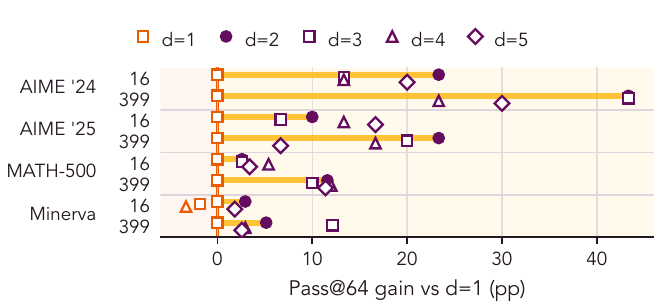}
\caption{Complete mathematics ladder at \pass{64}.}
\label{fig:app-math-xfer-full-pass64}
\end{subfigure}
\caption{Complete mathematics ladders behind
Figure~\ref{fig:response-diversity-math-transfer}, at \textbf{(a)} \pass{1} and
\textbf{(b)} \pass{64}. Each mark is one $d{=}2,\ldots,5$ score difference from $d{=}1$
in points, for one benchmark and pool (16 and 399). Here $d$ counts teacher sources
at a fixed SFT trajectory budget. Scores are measured after RL on
reasoning-gym tasks, at RL step 350, and neither SFT nor RL trains on the
four benchmarks. All 32 differences are positive at \pass{1}, and 30 of
32 are positive at \pass{64}. Marks left of zero favor $d{=}1$.}
\label{fig:app-response-diversity-math-full}
\end{figure}

The lead carries to standard mathematics benchmarks, AIME 2024, AIME 2025, MATH-500 and Minerva, which are not part of either training set. Evaluated on the same step-350 RL checkpoints, five teachers exceed one on all eight benchmark and pool combinations at both \pass{1} and \pass{64} (Figure~\ref{fig:response-diversity-math-transfer}). Across the complete sweep, all 32 comparisons with the single-teacher condition are positive at \pass{1}, and 30 of 32 are positive at \pass{64}. The two exceptions are Minerva in the 16-environment pool at three and four teachers, 1.8 and 3.3 points below one teacher at \pass{64} (Figure~\ref{fig:app-response-diversity-math-full}).
The Minerva reversal begins at \pass{8} for three teachers and \pass{4}
for four teachers, and persists at every larger reported budget. Teacher
count is nonmonotonic here too: the unweighted mean over the eight benchmark
and pool combinations at \pass{1}/\pass{64} is $30.36\%/59.04\%$ for two
teachers and $21.48\%/55.32\%$ for five. Each mathematics evaluation uses
64 draws per problem at temperature 1.0 and an 8,192-token cap; all
\pass{k} estimates for a condition reuse those draws.

\subsection{Teacher-source count after mathematics RL}
\label{app:dapo}

This sweep starts from the same Qwen3-1.7B teacher-source SFT checkpoints
as the mathematics-transfer sweep in Figure~\ref{fig:response-diversity-math-transfer},
trained by SFT on the 16- and 399-environment pools of Section~\ref{sec:modelcount} with
$d{=}1,\ldots,5$ teachers at an equal trajectory budget. RL then trains each condition with
GRPO on the DAPO-Math-17k mathematics set \citep{dapo2025}, with identical settings for every
condition, in place of RL on reasoning-gym tasks. Mathematics is out of distribution for the
SFT domain. We evaluate the RL
step-50 checkpoint on AIME 2024, AIME 2025, MATH-500, and Minerva with 64 samples per
problem, temperature 1.0, and an 8,192-token generation cap. Each AIME edition
has 30 problems, MATH-500 has 500, and Minerva has 272.

\begin{figure}[!htbp]
\centering
\begin{subfigure}[t]{\linewidth}
\centering
\includegraphics[width=4.4in]{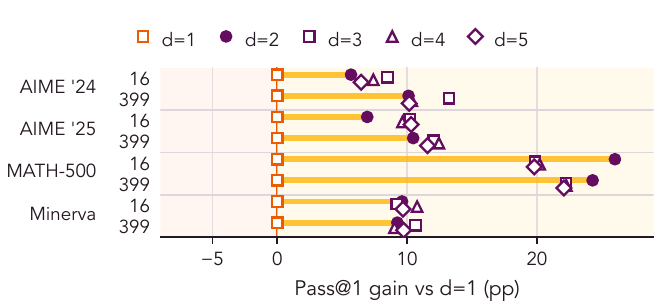}
\caption{Mathematics at \pass{1}.}
\label{fig:app-dapo-pass1}
\end{subfigure}
\par\medskip
\begin{subfigure}[t]{\linewidth}
\centering
\includegraphics[width=4.4in]{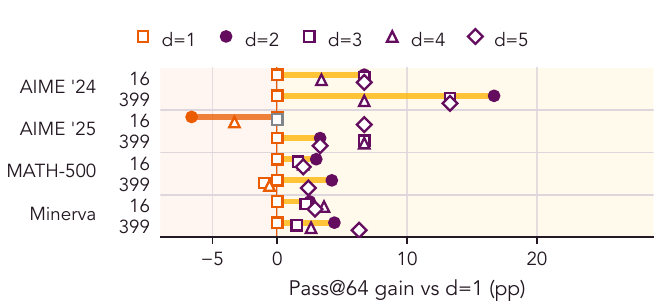}
\caption{Mathematics at \pass{64}.}
\label{fig:app-dapo-pass64}
\end{subfigure}
\caption{Qwen3-1.7B SFT conditions with $d{=}1,\ldots,5$ teacher sources at equal trajectory counts over environment pools 16 and 399, then identical GRPO on DAPO-Math-17k mathematics, outside the SFT domain. Each mark is the score difference from $d{=}1$ for one benchmark and pool at \textbf{(a)} \pass{1} and \textbf{(b)} \pass{64}.}
\label{fig:app-dapo-math}
\end{figure}

All 32 comparisons of a multi-teacher condition against $d{=}1$ (four
teacher counts, four benchmarks, two pools) are positive at \pass{1}, with a mean gain
of 12.8 points, and all 32 stay positive at every budget up to \pass{8}. At \pass{64},
27 are higher, one is unchanged, and four are lower, with a mean gain of 4.1 points
(Figure~\ref{fig:app-dapo-math}). The four \pass{64} losses are AIME 2025
in pool 16 with two and four teachers ($-6.67$ and $-3.33$ points), and
MATH-500 in pool 399 with three and four teachers ($-1.00$ and $-0.60$
points). AIME 2025 in pool 16 ties at three teachers. The MATH-500 losses
begin at \pass{16}, and the AIME 2025 losses at \pass{32}. Thus the
advantage extends to RL on mathematics for all comparisons through eight
samples, while larger sampling budgets include reversals. Means weight the
32 benchmark, pool, and teacher-count contrasts equally.

\subsection{Qwen3-4B mathematics evaluation}
\label{app:math-qwen4b-sources}

Table~\ref{tab:math-qwen4b-sources} gives the individual scores behind
Figure~\ref{fig:teacher-source-math-qwen4b}. The SFT data come from RLVE pools
of 16 or 399 environments. Both conditions use verified solutions to the
same prompt pool at equal total SFT trajectory counts, supplied by one or
twelve teachers from the ordered roster in Appendix~\ref{app:rosters}.
Both conditions receive the same SFT recipe and RL on DAPO-Math-17k,
and are evaluated at RL step 200.

\begin{table}[htbp]
\centering
\caption{Qwen3-4B mathematics scores in percent, higher is better, at saved
checkpoint step 200. Pool is the SFT environment pool.
Evaluation uses 64 samples per question, temperature 1.0, and an 8,192-token
generation cap. AIME has 30 questions per edition, MATH-500 has 500, and Minerva
has 272. There is one training run per condition. \pass{1} averages
correctness over the 64 draws; \pass{64} is the fraction of problems with
at least one correct draw.}
\label{tab:math-qwen4b-sources}
\small
\AppendixTableStyle
\begin{tabular}{lrrrrr}
\toprule
\rowcolor{AppendixTableHeader}
& & \multicolumn{2}{c}{\pass{1} (\%)} & \multicolumn{2}{c}{\pass{64} (\%)} \\
\cmidrule(lr){3-4}\cmidrule(lr){5-6}
\rowcolor{AppendixTableHeader}
Benchmark & Pool & 1 teacher & 12 teachers & 1 teacher & 12 teachers \\
\midrule
AIME 2024 & 16 & 16 & 22 & 43 & 50 \\
AIME 2024 & 399 & 16 & 19 & 47 & 53 \\
AIME 2025 & 16 & 9.5 & 17 & 43 & 53 \\
AIME 2025 & 399 & 11 & 19 & 40 & 53 \\
MATH-500 & 16 & 34 & 65 & 81 & 83 \\
MATH-500 & 399 & 46 & 71 & 82 & 89 \\
Minerva & 16 & 14 & 27 & 54 & 65 \\
Minerva & 399 & 17 & 31 & 53 & 61 \\
\bottomrule
\end{tabular}
\end{table}

Twelve teachers score higher than one teacher in all 16 cells of
Table~\ref{tab:math-qwen4b-sources}. The largest gain is 30.9 points of \pass{1}
on MATH-500 in pool 16, from 34.14\% to 65.08\%. The same 64 samples per question
also give \pass{k} at $k \in \{2, 4, 8, 16, 32\}$. Across the seven budgets from
$k{=}1$ to $k{=}64$, twelve teachers lead in 54 of the 56 benchmark, pool, and
budget cells. The two exceptions are AIME 2025 in pool 16 at \pass{8}, $29.89\%$
against $30.57\%$ for one teacher, and at \pass{16}, $35.52\%$ against
$35.58\%$. A separate evaluation of the same checkpoints with a 32,768-token cap
gives the same direction in 15 of the 16 \pass{1} and \pass{64} cells. The
exception is AIME 2024 \pass{64} in pool 16, where one teacher solves 16 of the
30 problems and twelve teachers solve 14. The two cap evaluations use
separate stochastic draws from the same checkpoints. At the 8,192-token cap,
all displayed \pass{1} and \pass{64} comparisons favor twelve teachers;
intermediate budgets and the separate longer-cap evaluation include reversals.

\section{Supplementary Real-Data Details}
\label{app:secondary}

\subsection{Dolci-Think selection for reward diagnostics}
\label{app:dolci-checkpoints}

The pre-RL reward-signal diagnostic in Section~\ref{sec:analysis} uses two
100{,}000-row SFT datasets selected from the same pool of 245{,}571 verified
Dolci-Think candidates. The twelve-model roster in Appendix~\ref{app:rosters}
generates complete solutions to the shared prompts, and the answer checker
retains accepted solutions. Both selections allocate 25{,}000 rows each to
mathematics, science, verified synthetic tasks, and instruction following.
Each reasoning-topology fingerprint concatenates 137 continuous trace
statistics, 430 tree-traversal features, and 64 binary pattern indicators,
giving 631 raw features. The construction projects these fingerprints to
96 dimensions with seed 42 and uses 40{,}000 clusters with selection seed 42.
The diverse condition allocates each domain's budget across fingerprint
clusters in proportion to cluster size and runs farthest-point selection inside
each cluster. The similar condition keeps the rows nearest each domain's centroid.
OLMo3-7B receives the same SFT recipe in both conditions: batch size 32,
learning rate $10^{-5}$, and sequence-length limit 16{,}384. The diagnostic
uses the SFT step-3{,}120 checkpoints.

The diagnostic compares the resulting checkpoints at the end of SFT, before
RL. Each checkpoint supplies eight responses at temperature 1.0 and top-$p=1.0$,
with a 30{,}720-token generation limit, to the same 64 mathematics prompts
drawn from the shared Dolci-RL-Zero-Mix training pool.
Appendix~\ref{app:metrics} defines the mixed-reward statistic.

\subsection{Released-corpus construction and training}
\label{app:realdata-construction}

The released corpora follow the shared procedure of
Appendix~\ref{app:selection-method}. This subsection gives their settings.

\paragraph{Pool and fingerprint.}
Candidates are the released single-turn conversations with complete reasoning
spans, used as released. Domain labels come from the corpus metadata. The
common Nemotron-Cascade 2 pool also enforces the 50{,}000-character limit of
Appendix~\ref{app:realdata-baselines}. The fingerprint uses all five blocks:
137 continuous, 430 tree, 64 pattern, 83 conversation-level and 1{,}024
signed-hash features, so $D=1{,}738$. Each coordinate is standardized over the
corpus's normalized input conversations, before the Nemotron-Cascade 2
eligibility mask, with the variance floored at $10^{-8}$. A seed-42 Gaussian
projection maps the vectors to 96 dimensions. Both conditions and the topology
baseline use this same feature map.

\paragraph{Quotas and clustering.}
OpenThoughts3 and INTELLECT-3 each split their 100{,}000 rows into 33{,}334
mathematics, 33{,}333 code and 33{,}333 science rows (the selector calls the
science domain \texttt{stem}). Nemotron-Cascade 2 allocates all 100{,}000 rows
to mathematics. The shared total of $K=40{,}000$ clusters gives 13{,}334
mathematics and 13{,}333 each for code and science on OpenThoughts3 and
INTELLECT-3, and 40{,}000 mathematics clusters on Nemotron-Cascade 2.
Mini-batch $k$-means uses seed 42, three initializations, at most 50 iterations
and minibatches of $\min(65{,}536,|E_d|)$ rows.

\paragraph{SFT and RL recipe.}
The student is OLMo3-7B with its Think chat template. SFT is full-parameter,
one epoch, global batch size 32, learning rate $10^{-5}$ and sequence-length
limit 16{,}384. GRPO uses the shared Dolci-RL-Zero-Mix mixture of mathematics,
code, instruction following and science, with 128 prompts per step, $G=8$
responses per prompt, minibatch size 128, learning rate $10^{-6}$, KL
coefficient $\beta=0.001$, no entropy term, and prompt and response limits of
4{,}096 and 16{,}384 tokens, over a 64-step schedule. Every reported comparison
evaluates both conditions at the same RL checkpoint step, specified with its
results.

\paragraph{Released-corpus evaluation.}
Evaluation uses the Think template, temperature $0.7$, top-$p=0.95$,
a maximum of 30{,}720 generated tokens and a 32{,}768-token context, with the
\acc{} and \pass{k} definitions of Appendix~\ref{app:selection-method}.
Appendices~\ref{app:realdata-per-benchmark}
and~\ref{app:realdata-relative-benchmarks} specify the benchmark sets, sample
counts and aggregation for the Diverse and Similar comparison and for the
selection-baseline comparison. The separate pre-RL Dolci diagnostic keeps its
temperature-$1.0$ setting stated above.

\Needspace{8\baselineskip}
\subsection{Released-corpus results by benchmark}
\label{app:realdata-per-benchmark}

\begin{figure}[!htbp]
\centering
\includegraphics[width=5.5in]{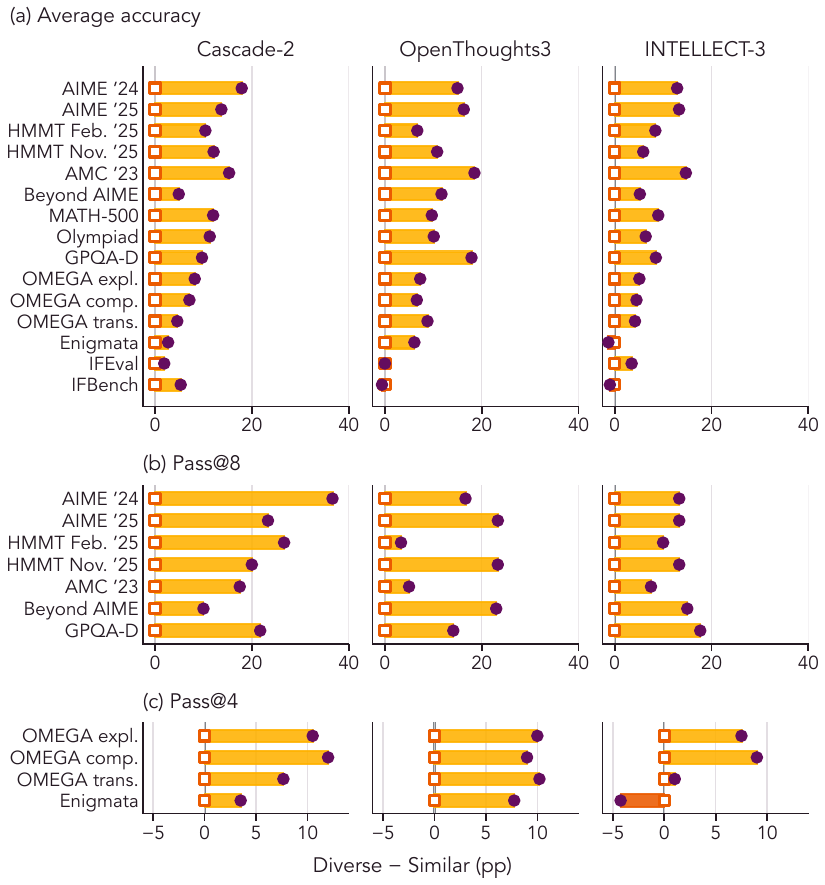}
\caption{Three released corpora, with both conditions at the same RL step.
Open squares mark Similar at zero, filled circles Diverse, and bars their signed
difference in points. (a) Mean sampled accuracy, with strict prompt accuracy
for IFEval and IFBench. (b) \pass{8} for the seven eight-response benchmarks.
(c) \pass{4} for OMEGA and Enigmata. All rows use RL step 8
for Nemotron-Cascade 2 and OpenThoughts3, and step 16 for INTELLECT-3.
On INTELLECT-3 Enigmata, Diverse reaches $1.75\%$ \pass{4},
compared with $6.00\%$ for Similar.}
\label{fig:realdata-corpora-intellect3}
\label{fig:realdata-transfer}
\end{figure}

Figure~\ref{fig:realdata-corpora-intellect3} compares the Diverse
and Similar conditions on every reported benchmark at the same RL step.
The caption identifies the checkpoint used for each corpus.

\paragraph{Benchmarks.}
\label{app:realdata-benchmarks}
The mathematics evaluations use AIME 2024 and 2025 and AMC 2023
\citep{maaamc}, HMMT February and November 2025
\citep{hmmt2025feb,hmmt2025nov}. They also include Beyond AIME \citep{bytedance2025beyondaime},
the MATH-500 subset of MATH \citep{hendrycks2021math,lightman2024verify}, and
OlympiadBench \citep{he2024olympiadbench}. Science, puzzle, and instruction-following
evaluations use GPQA-Diamond \citep{rein2024gpqa}, Enigmata
\citep{chen2025enigmata}, IFEval \citep{zhou2023ifeval}, and IFBench
\citep{pyatkin2025ifbench}. Held-out mathematics also includes the
three OMEGA splits, explorative, compositional and transformative
\citep{sun2025omega}. Each experiment reports its evaluated subset of these
benchmarks.

Mean sampled accuracy averages correctness over responses and then questions.
The seven eight-response benchmarks are AIME 2024 and 2025, HMMT February and
November 2025, AMC 2023, Beyond AIME, and GPQA-Diamond. MATH-500, OlympiadBench,
the three OMEGA splits, and Enigmata use four responses per question. IFEval
and IFBench report strict prompt-level instruction-following accuracy from one
response per question. Coverage at budget $k$ is the fraction of questions
with at least one correct answer among $k$ responses. Selection-baseline
evaluations use eight responses per question
(Appendix~\ref{app:realdata-relative-benchmarks}).

The gains are not uniform across capabilities. IFBench accuracy decreases by
0.6 points on OpenThoughts3 and 1.0 on INTELLECT-3, while OpenThoughts3 IFEval
ties. INTELLECT-3 Enigmata also favors Similar, by 1.3 points in mean accuracy
and 4.25 points in \pass{4}. These exceptions coexist with the mathematics,
OMEGA and GPQA-Diamond gains reported in Section~\ref{sec:realdata}. Each
condition has one training run; the observed differences do not measure
variation across training seeds.

\subsection{Released-corpus selection baselines}
\label{app:realdata-baselines}

\paragraph{Baselines.}
\label{app:baselines}
Every baseline selects the same budget of $k{=}100{,}000$ rows from the
same pool as our method, under the same per-domain quotas (one third each of
math, code and science on two pools, and all math on the third), with a fixed
seed. The matched selections receive the same one-epoch
SFT and 64-step GRPO recipe (Appendix~\ref{app:realdata-construction}). The
baselines differ only in the criterion that picks the rows.
For Nemotron-Cascade 2, all selectors use the same 1{,}353{,}746-row eligible
pool: from 2{,}142{,}332 normalized rows, we keep conversations containing at
most 50{,}000 characters across all messages.

\textbf{Random.} Uniform sampling without replacement within the domain quotas
\citep{diddee2024chasing}.

\textbf{Topology baseline.} This baseline uses our
96-dimensional topology fingerprints with a simpler rule than our method:
plain farthest-point selection on OpenThoughts3, and random sampling within
fingerprint clusters in proportion to cluster size on INTELLECT-3 and
Nemotron-Cascade 2. Our method allocates the budget across
fingerprint clusters in proportion to cluster size and then runs
farthest-point selection inside each cluster (Table~\ref{tab:config}). On
OpenThoughts3 the baseline runs greedy farthest-point sampling
\citep{eldar1997farthest,coreset2018} per domain on unit-normalized vectors:
each pick maximizes its distance to the nearest selected row. The exact
implementation handles pools of up to 600{,}000 rows, which covers
OpenThoughts3 (about 456{,}000 rows). On INTELLECT-3 and Nemotron-Cascade 2,
the eligible pools exceed this threshold. The baseline instead samples
randomly within each domain, allocating its quota in proportion to eligible
rows' membership in 200 global topology-fingerprint clusters
\citep{cochran1977sampling}.

\textbf{Embedding farthest-point sampling.} The same farthest-point
rule applied to general-purpose sentence embeddings: each conversation, with
its message contents concatenated and truncated to 8{,}192 tokens, is embedded
with Qwen3-Embedding-8B \citep{zhang2025qwen3embedding} using the model's
native last-token pooling and L2 normalization.

\textbf{Gradient-diversity selection (G-Vendi proxy).} A forward pass through
OLMo3-7B produces the surrogate
$\operatorname{mean}_t W^\top(p_t-y_t)$ over 128 token positions, using a
top-512-vocabulary approximation. Here $W$ is the output projection, $p_t$ the
predicted token distribution, and $y_t$ the one-hot target. This
gives a 4{,}096-dimensional vector that is L2-normalized. The baseline
selects by greedy farthest-point sampling in this space within each domain.
This is a gradient-diversity proxy \citep{jung2025prismatic,friedman2023vendi};
it does not compute full parameter gradients or optimize the Vendi score.

\textbf{Lexical farthest-point sampling.} Farthest-point selection using
OLMo3 tokenizer unigrams and
bigrams of the concatenated message contents, hashed into $2^{21}$ buckets
\citep{weinberger2009hashing}. Sublinear term frequency and
smoothed inverse document frequency (TF-IDF) \citep{salton1988term} weight the
features, with IDF estimated from a fixed 5\% sample of the candidate pool. A
seeded $\pm1$ random projection \citep{achlioptas2003database}
reduces the vectors to 1{,}024 dimensions, followed by L2 normalization.
For each farthest-point baseline, a candidate's selection score is its distance
to the nearest row already selected in that baseline's feature space, and the
next pick maximizes this score. Distances are squared Euclidean, and the first
row is chosen randomly with selection seed 42. This random initialization is
specific to the baselines. Our selector starts from the centroid-nearest row.

\textbf{Similar condition.} The low-diversity end of the comparison
is the Similar condition, which keeps the rows nearest each domain's centroid
and so draws from a dense region of topology-fingerprint space, as in
Section~\ref{sec:measure}. Appendix~\ref{app:realdata-per-benchmark} compares
it with the diverse selection.

On OpenThoughts3 we train the random, topology, gradient-diversity, embedding
and lexical baselines, on INTELLECT-3 the random, topology, gradient-diversity
and embedding baselines, and on Nemotron-Cascade 2 the random, topology and
gradient-diversity baselines.

Figure~\ref{fig:realdata-baseline-gains} summarizes relative gains in the
mean benchmark score over these baselines. Every selection-baseline comparison
on OpenThoughts3, INTELLECT-3, and Nemotron-Cascade 2 uses the final RL
checkpoint, step 64, for both conditions. The evaluations use eight responses
per question; mean sampled accuracy and \pass{8} are defined in
Appendix~\ref{app:metrics}. The OpenThoughts3 aggregate covers six benchmarks,
excluding HMMT February 2025; the other two corpora cover seven.
Appendix~\ref{app:realdata-relative-benchmarks} specifies the benchmark sets,
defines the aggregation, and reports every benchmark-level gain.
The Diverse--Similar comparisons in Appendix~\ref{app:realdata-per-benchmark}
also use the same RL step for both conditions: step 8 on OpenThoughts3 and
Nemotron-Cascade 2, and step 16 on INTELLECT-3. Each arm has one training run, so
these comparisons do not measure variation across training seeds.

\paragraph{Selection cost.}
\label{app:selection-cost}
For Nemotron-Cascade 2, preprocessing, fingerprinting, and selection took
about three hours on one CPU node, with no GPU computation. Fingerprinting
processes the 2{,}142{,}332 normalized input rows; selection then keeps
100{,}000 rows from the shared 1{,}353{,}746-row eligible pool described in
Appendix~\ref{app:realdata-baselines}.
On the roughly two-million-row corpora, the gradient-diversity baseline used
64 to 92 H100 GPU-hours and the embedding baseline used 175 to 232 H100
GPU-hours. These methods require an OLMo3-7B or Qwen3-Embedding-8B forward
pass over the candidate rows, respectively. The CPU figure is elapsed time;
the GPU figures report GPU-hours across the corresponding jobs.

\FloatBarrier
\subsection{Benchmark-level relative gains}
\label{app:realdata-relative-benchmarks}

\begin{figure}[!htbp]
\centering
\includegraphics[width=5.5in]{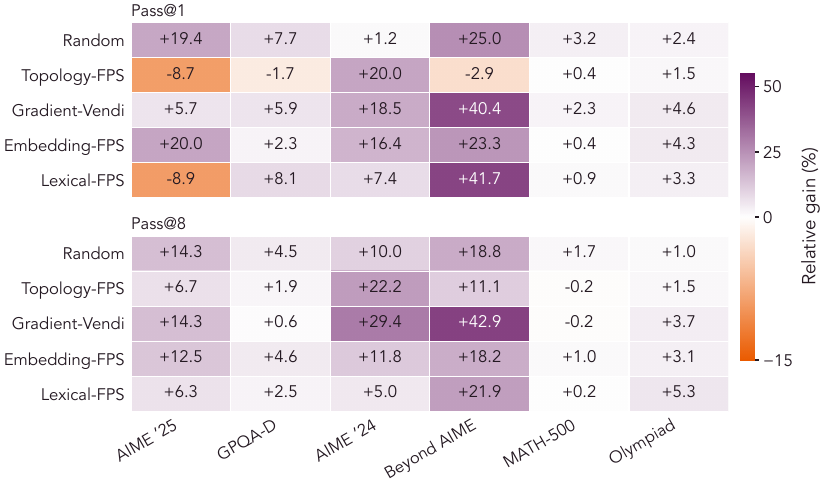}
\caption{OpenThoughts3: relative gains over each selection baseline on
individual benchmarks. Purple indicates positive gains, orange losses, and
white zero. The \pass{1} and \pass{8} panels show every reported
benchmark cell.}
\label{fig:relative-benchmarks-ot3}
\end{figure}

\begin{figure}[!htbp]
\centering
\includegraphics[width=5.5in]{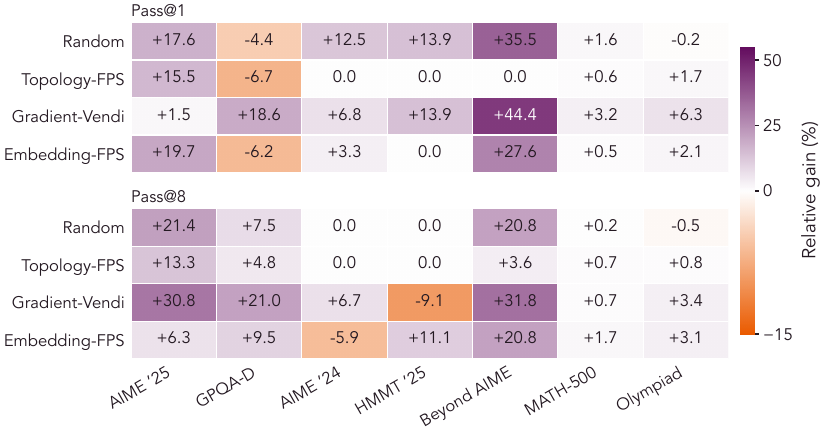}
\caption{INTELLECT-3: benchmark-level relative gains over each selection
baseline, on the same scale as Figure~\ref{fig:relative-benchmarks-ot3}.
Ties and losses remain visible alongside improvements.}
\label{fig:relative-benchmarks-int3}
\end{figure}

\begin{figure}[!htbp]
\centering
\includegraphics[width=5.5in]{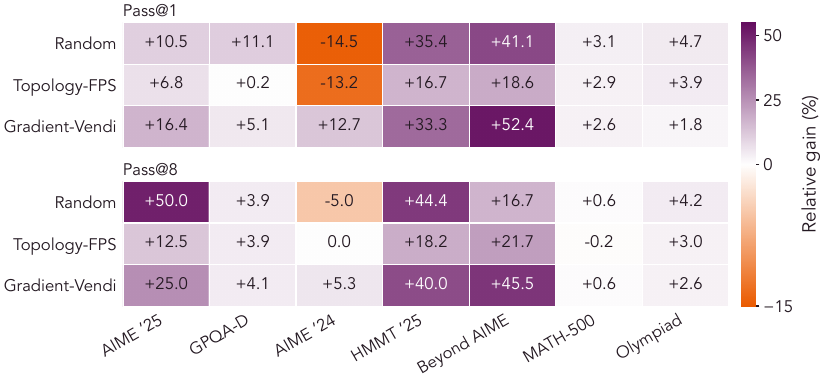}
\caption{Nemotron-Cascade 2: benchmark-level relative gains over each selection
baseline, on the same scale as Figure~\ref{fig:relative-benchmarks-ot3}.
All three evaluated approaches are shown.}
\label{fig:relative-benchmarks-casc2}
\end{figure}

Figures~\ref{fig:relative-benchmarks-ot3}--\ref{fig:relative-benchmarks-casc2}
show the individual benchmark gains behind the aggregate comparisons in
Figure~\ref{fig:realdata-baseline-gains}. Positive values favor our selection;
negative values favor the named baseline. All panels share the same color
scale. For benchmark $b$, let $s_{D,b}$ and $s_{B,b}$ be the scores of the
diverse selection and the named baseline under the matched protocol. A cell
reports $100(s_{D,b}/s_{B,b}-1)$, a relative percentage rather than a difference
in percentage points. The panels labeled \pass{1} use mean sampled accuracy
over eight responses per question, and \pass{8} uses any-correct coverage over
those responses.

The main figure instead reports the relative change in the equally weighted
mean benchmark score:
\[
  100\left(\frac{\bar s_D}{\bar s_B}-1\right),
  \qquad
  \bar s_A=\frac{1}{|\mathcal B|}\sum_{b\in\mathcal B}s_{A,b}.
\]
For OpenThoughts3, $\mathcal B$ contains AIME 2025, GPQA-Diamond, AIME 2024,
Beyond AIME, MATH-500 and OlympiadBench. INTELLECT-3 and Nemotron-Cascade 2
add HMMT February 2025. Thus the aggregate weights benchmarks equally before
taking the relative change; it is neither a question-weighted pooled score nor
an average of the relative percentages printed in these cells. Display rounding
is applied after aggregation.
The row label Topology-FPS marks the topology baseline, which runs plain
farthest-point selection on OpenThoughts3 and samples within
topology-fingerprint clusters in proportion to cluster size on the larger
INTELLECT-3 and Nemotron-Cascade 2 pools, whose sizes exceed the 600{,}000 rows
its exact farthest-point implementation handles. The row label Gradient-Vendi
marks the gradient-diversity baseline, which runs farthest-point selection on
gradient features as a proxy for G-Vendi. The row labels Embedding-FPS and
Lexical-FPS mark farthest-point selection on sentence embeddings and on
lexical features. All baselines are defined in
Appendix~\ref{app:baselines}.

\FloatBarrier

\section{Extended Results}
\label{app:extended}

This section gives additional coverage results for RLVE, OMEGA, Sokoban and
program simulation. Construction sweeps and real-data results appear above.

\subsection{RLVE selection budgets (Qwen3)}
\label{app:rlve-qwen-budget}
\begin{figure}[htbp]
\centering
\includegraphics[width=4.4in]{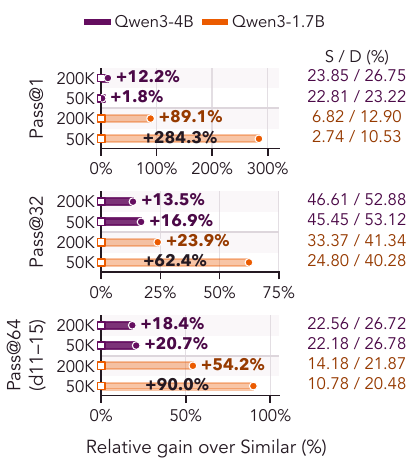}
\caption{Qwen3 students on RLVE after the same RL: relative gain of
Diverse over Similar SFT. Diverse leads on every metric at both model sizes and
both selection budgets. The right column gives the Similar and Diverse scores
in percent. \pass{1} denotes mean sampled accuracy, and \pass{32} and \pass{64}
denote sampled coverage.}
\label{fig:rlve-qwen-budget}
\end{figure}

Figure~\ref{fig:rlve-qwen-budget} gives the relative gain of Diverse over
Similar SFT for both Qwen3 students at both selection budgets after the same RL.

\paragraph{Lexical diversity among correct completions.}
\label{app:answer-diversity}
Figure~\ref{fig:rlve-answer-diversity} uses the checkpoints after 75 RL steps
and 32 responses per question on 3,822 RLVE questions, with temperature 0.7,
top-$p=0.95$, and a 4,096-model-token response cap. A question is eligible
when both conditions produce at least two correct completions long enough for
the chosen prefix. We match questions by environment, generator seed and
difficulty. Prefixes contain the first 128 or 512 whitespace-delimited tokens,
preserving case and markup. For two prefix bigram sets $A$ and $B$, their
Jaccard distance is $1-|A\cap B|/|A\cup B|$. We average this distance over
all unordered pairs within a question, then equally over the matched questions.
Thus each question contributes the expected distance between two uniformly
selected eligible correct completions; having more successful samples does
not give it more weight.

\begin{table}[!htbp]
\caption{Mean pairwise bigram Jaccard distance among correct completions.
Prefix lengths are whitespace-token counts. Confidence limits are percentages
for the relative gain, from paired resampling of evaluation environments.}
\label{tab:answer-diversity}
\centering
\small
\setlength{\tabcolsep}{5pt}
\AppendixTableStyle
\begin{tabular}{lrrrrl}
\toprule
\rowcolor{AppendixTableHeader}
Student & Prefix & Questions & Diverse & Similar & Relative gain (95\% CI) \\
\midrule
Qwen3-4B & 128 & 861 & 0.74 & 0.64 & $17\%$ ($14$, $19$) \\
 & 512 & 675 & 0.82 & 0.77 & $6.0\%$ ($5.2$, $6.8$) \\
Qwen3-1.7B & 128 & 1,327 & 0.57 & 0.49 & $15\%$ ($12$, $18$) \\
 & 512 & 1,258 & 0.74 & 0.71 & $4.0\%$ ($3.3$, $4.7$) \\
\bottomrule
\end{tabular}

\end{table}

The 95\% intervals use a paired percentile bootstrap over environment IDs
with 20,000 resamples and seed 20260924. Each sampled environment retains all
its eligible questions, and each resample recomputes the question-weighted
means and their relative difference. These intervals describe evaluation-set
variation, not training-seed uncertainty. The longer-prefix gains are smaller,
and each prefix length has a different eligible question set. This measurement
supports greater lexical variety among correct completions; it is not a direct
count of semantic reasoning routes.

\subsection{RLVE per-difficulty (Qwen3-4B-Base)}

A second Qwen3-4B-Base pair at 200{,}000 SFT rows is split here by generator difficulty. Both of its conditions are selected from one shared RLVE pool with the same procedure, and both receive the same 75-step GRPO run. The diverse condition leads at all fifteen difficulties, by 1.5 to 7.4 points of mean sampled accuracy and by 3.4 to 10.4 points of coverage, measured as \pass{32} at difficulties 1 to 10 and \pass{64} at 11 to 15. Difficulties 11 to 15 lie above the range used in SFT and RL, so the lead reaches problems harder than either stage practiced.

\begin{figure}[!htbp]
\centering
\begin{subfigure}[t]{\linewidth}
\centering
\includegraphics[width=4.4in]{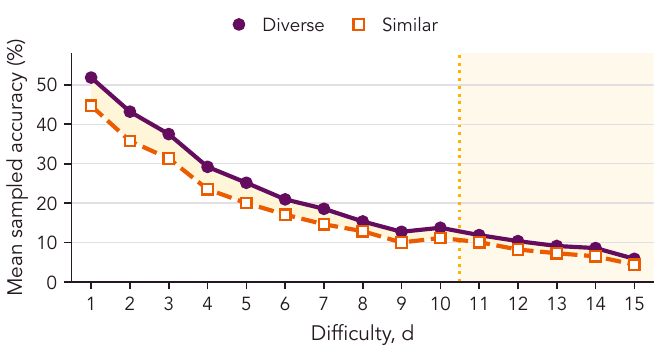}
\caption{Mean sampled accuracy.}
\label{fig:app-rlve-perdiff-accuracy}
\end{subfigure}
\begin{subfigure}[t]{\linewidth}
\centering
\includegraphics[width=4.4in]{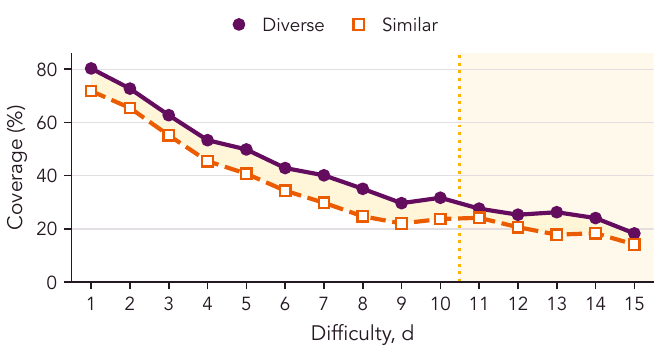}
\caption{Coverage.}
\label{fig:app-rlve-perdiff-coverage}
\end{subfigure}
\caption{\textbf{(a)}~Mean sampled accuracy and \textbf{(b)}~coverage in percent by RLVE
difficulty, after identical RL from diverse (purple circles) and similar (orange squares)
SFT. Difficulties 1 to 10 use 32 samples. The shaded range, 11 to 15, uses 64.
Qwen3-4B-Base with 200{,}000 SFT rows selected from the shared pool.
Mean sampled accuracy uses the scored question set, and coverage uses its
programmatic-reference-answer subset. This pair is separate from the
Qwen3-4B 200{,}000-row pair in Figure~\ref{fig:rlve-qwen-budget}.}
\label{fig:app-rlve-perdiff}
\end{figure}

\subsection{OMEGA held-out mathematics}
\label{sec:sftcompound}

OMEGA tests held-out mathematics in
natural language, split into explorative, compositional, and transformative
problems \citep{sun2025omega}. It separates an in-distribution pool from
out-of-distribution (OOD) pools whose problems combine or transform skills. Candidate
solutions to OMEGA training prompts are generated by open reasoning models
(Appendix~\ref{app:rosters}) and kept when the benchmark's answer checker
accepts them. From this one accepted pool we select 50{,}000 diverse and
50{,}000 similar traces with the strategy-step fingerprint and train OLMo3-7B
by SFT on each. Both then run the same 75-step GRPO on 4{,}670 prompts drawn
from the base and OOD pools, and evaluation uses 500 held-out OOD prompts at up
to 64 samples.
The diverse selection reaches
$45.4\%$ \pass{64}, compared with $39.4\%$ for the similar selection
(Figure~\ref{fig:omega-models}). The coarser roster contrast on a separate
300-problem subset of the same held-out OOD pool, across three Qwen3
capacities, is reported below.

\paragraph{Student capacity.}
\leavevmode OMEGA also tests the teacher-source contrast across
student capacity, with three Qwen3 base models as students. Unlike the sweeps
of Section~\ref{sec:modelcount}, this comparison takes its SFT prompts from
OMEGA. For each student, a
corpus written by a multi-model roster is compared with a corpus of the same
number of solutions written by Qwen3-4B alone (Appendix~\ref{app:rosters}). RL
trains on part of OMEGA's out-of-distribution set, and evaluation uses 300
separate held-out prompts from that set (Table~\ref{tab:config}).
The multi-model corpus leads at \pass{64} for all three students,
most at the smallest capacity. At 4B it solves $33.3\%$ of held-out prompts against $21.7\%$, and at 14B it
leads by 1.3 points, or four prompts (Figure~\ref{fig:omega-models}).

\subsection{Diagnostics where the route is executable}
\label{sec:olmo-explicit-routes}
\begin{figure}[htbp]
\centering
\begin{subfigure}[t]{2.35in}
\centering
\includegraphics[width=2.35in]{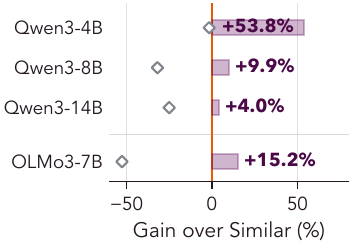}
\caption{OMEGA relative gains.}
\label{fig:omega-models}
\end{subfigure}\hfill
\begin{subfigure}[t]{1.50in}
\centering
\includegraphics[width=1.50in]{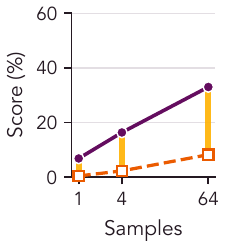}
\caption{Program simulation.}
\label{fig:olmo-program}
\end{subfigure}\hfill
\begin{subfigure}[t]{1.50in}
\centering
\includegraphics[width=1.50in]{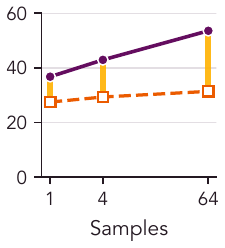}
\caption{Sokoban.}
\label{fig:olmo-sokoban}
\end{subfigure}
\caption{Post-RL mathematical and executable-route comparisons.
Diverse leads Similar at \pass{64} in every panel, and in (b,c) the gap grows with $k$.
\textbf{(a)}~Relative OMEGA \pass{64} gains, $100(\mathrm{condition}-\mathrm{Similar})/\mathrm{Similar}$.
Purple bars show Diverse, the orange zero line denotes Similar, and gray diamonds
show direct RL without SFT on the same relative scale.
For Qwen3, Diverse and Similar denote multi-model and single-model corpora;
for OLMo3-7B, they denote topology-selected subsets of one shared pool.
\textbf{(b,c)}~OLMo3-7B on program simulation and Sokoban at $k=1,4,64$.
In both, $k=1$ is sampled \pass{1}.
Both panels estimate \pass{4} from 64 samples per problem and report
empirical coverage at $k=64$ (Appendix~\ref{app:metrics}). Scores are percentages. In
(b), Diverse is the multi-model corpus and Similar the single-model corpus of
equal size. In (c), both are selected from one pool.
Purple circles mark Diverse and orange open squares Similar.
Both conditions in each comparison use RL step 75. One run per condition.}
\label{fig:olmo-explicit-routes}
\end{figure}

A Sokoban solution is a replayable state-action
path \citep{wang2025ragen} and a program-simulation solution is a trace through
rewrite systems adapted from RLVE's A::B environment \citep{zeng2026rlve}, so in
both the step structure of a route can be checked directly. Program simulation
also holds the task family fixed, so a gain there cannot come from broader task
coverage.
Sokoban selects both conditions from one shared pool of verified traces with
the shared topology-based procedure (Appendices~\ref{app:selection-method}
and~\ref{app:sokoban-construction}). Program
simulation contrasts two corpora of equal size,
one written by several models and one by a single model. The student is
OLMo3-7B under the same 75-step GRPO (Table~\ref{tab:config}). At 64 samples the
diverse condition solves 53.6\% of held-out Sokoban boards against 31.4\% for
the similar condition (Figure~\ref{fig:olmo-sokoban}). On program simulation the diverse
condition solves 33.0\% of prompts against 8.2\%
(Figure~\ref{fig:olmo-program}). In both, the gap grows with the number of
samples.

\clearpage
\section{Per-Testbed Configuration}
\label{app:config}

\begingroup
\small
\setlength{\tabcolsep}{6pt}
\setlength{\LTcapwidth}{\linewidth}
\renewcommand{\arraystretch}{1.18}
\arrayrulecolor{AppendixTablePurple!45!white}
\begin{longtable}{@{}>{\bfseries\color{AppendixTableInk}\raggedright\arraybackslash}p{0.21\linewidth} >{\raggedright\arraybackslash}p{\dimexpr0.79\linewidth-2\tabcolsep\relax}@{}}
\caption{Setup for the RLVE, OMEGA, Sokoban, program-simulation, real-data and single-teacher testbeds. Each block gives candidate generation and verification, the diverse and similar conditions, the evaluation, and the SFT-to-GRPO training recipe.}
\label{tab:config}\\
\toprule
\endfirsthead
\multicolumn{2}{@{}l}{\tablename\ \thetable\ (continued)}\\
\toprule
\endhead
\bottomrule
\endfoot
\bottomrule
\endlastfoot
\rowcolor{AppendixTableHeader}
\multicolumn{2}{@{}l}{\bfseries\color{AppendixTableInk} RLVE}\\*
Candidates and verification & Procedurally generated reasoning-gym environments, each with a rule-based verifier. Each trace gets a 1{,}373-dimensional lexical-topological fingerprint with 159 continuous, 1{,}150 topology and 64 pattern features.\\*
Selection conditions & Similar sets use nearest-centroid selection. Diverse sets cluster the fingerprints and use greedy farthest-point selection inside each cluster, with budgets in proportion to cluster size. At both Qwen3 sizes, both conditions select from the same shared pool at 50{,}000 and 200{,}000 rows. A second Qwen3-4B pair at 200{,}000 rows appears in Figure~\ref{fig:app-rlve-perdiff}. Both OLMo3-7B conditions retain 50{,}000 rows from a shared 64-environment pool.\\*
Evaluation & One fixed evaluation set spanning 384 environments and difficulties 1 to 15. Qwen3 runs report \pass{32} at difficulties 1 to 10 and \pass{64} at 11 to 15, over the 3{,}287 and 1{,}587 questions with a programmatic reference answer. The OLMo3-7B run reports \pass{8} and \pass{32} over 5{,}682 questions, 937 from the 63 SFT environments in the set and 4{,}745 from the 321 environments held out from SFT. The shared RL pool spans all 384 environments. Metrics and diagnostic caps are given in Appendix~\ref{app:metrics}.\\*
Training recipe & Full-parameter SFT for one epoch, learning rate $10^{-5}$, 8{,}192-token sequences. GRPO uses difficulties 1 to 10 for both Qwen3 and OLMo3-7B: 75 steps, 8 rollouts, actor learning rate $10^{-6}$, KL coefficient $0.001$, prompt 4{,}096 and response 16{,}384 tokens, at 128 prompts per step.\\
\addlinespace[6pt]
\rowcolor{AppendixTableHeader}
\multicolumn{2}{@{}l}{\bfseries\color{AppendixTableInk} OMEGA}\\*
Candidates and verification & Candidates are kept when the OMEGA answer checker accepts them. Single-model traces come from Qwen3-4B. Multi-model traces continue each response across a roster of open reasoning models and are kept in English (Appendix~\ref{app:rosters}). Topology selection uses a 323-dimensional strategy-step fingerprint over an accepted pool of 575{,}699 traces, 416{,}727 multi-model and 158{,}972 single-model.\\*
Selection conditions & The Qwen3 runs compare the multi-model corpus with the single-model corpus. The OLMo3-7B run keeps 50{,}000 diverse traces by clustered farthest-point selection and 50{,}000 similar traces by nearest-centroid selection.\\*
Evaluation & RL trains on 4{,}670 prompts, 600 in-distribution and 4{,}070 from the out-of-distribution pool. Evaluation uses 592 held-out out-of-distribution problems that share no problem with RL, read on deterministic subsets of 300 problems for the Qwen3 runs and 500 for OLMo3-7B, with 64 samples per problem.\\*
Training recipe & Qwen3 base models get 300 full-parameter SFT steps at batch 32 and learning rate $10^{-5}$, then 75 GRPO steps with 8 rollouts, actor learning rate $10^{-6}$, KL coefficient $0.001$, and a 2{,}048-token prompt and response cap. OLMo3-7B gets one full SFT epoch, then 75 GRPO steps with 8 rollouts, actor learning rate $10^{-6}$, prompt 4{,}096 and response 12{,}288 tokens.\\
\addlinespace[6pt]
\rowcolor{AppendixTableHeader}
\multicolumn{2}{@{}l}{\bfseries\color{AppendixTableInk} Real data}\\*
Candidates and verification & Dolci-Think prompts \citep{olmo3} with candidates generated by twelve open reasoning models and kept when the answer checker accepts them (245{,}571 verified rows); its fingerprint has 631 raw features (Appendix~\ref{app:dolci-checkpoints}). OpenThoughts3 \citep{guha2025openthoughts}, INTELLECT-3 \citep{intellect3} and Nemotron-Cascade 2 \citep{nemotroncascade2} contribute released single-turn solutions with complete reasoning spans and use 1{,}738 raw features. Both feature maps project to 96 dimensions (Appendix~\ref{app:realdata-construction}).\\*
Selection conditions & Both conditions select 100{,}000 rows from one shared pool with identical domain quotas. Diverse selection gives each fingerprint cluster a budget in proportion to its size and runs farthest-point selection inside each cluster. Similar takes the rows nearest each domain's centroid. Appendix~\ref{app:realdata-construction} gives the released-corpus quotas and clustering settings.\\*
Evaluation & The fifteen benchmarks for the released corpora are AIME 2024 and 2025, HMMT February and November 2025, AMC 2023, Beyond AIME, MATH-500, OlympiadBench, GPQA-Diamond, the three OMEGA splits, Enigmata, IFEval and IFBench. The released-corpus comparisons use eight draws on seven benchmarks, four on MATH-500, OlympiadBench, OMEGA and Enigmata, and one on IFEval and IFBench (Appendix~\ref{app:realdata-per-benchmark}). Selection-baseline comparisons use eight draws per question, over the benchmark sets in Appendix~\ref{app:realdata-relative-benchmarks}. The Dolci-Think pre-RL diagnostic uses eight draws on 64 mathematics training prompts (Appendix~\ref{app:dolci-checkpoints}).\\*
Training recipe & Released corpora: OLMo3-7B, one SFT epoch, batch 32, learning rate $10^{-5}$, 16{,}384-token sequences. The 64-update GRPO schedule uses 128 prompts per batch, eight rollouts, learning rate $10^{-6}$ and KL coefficient $0.001$ on Dolci-RL-Zero-Mix (Appendix~\ref{app:realdata-construction}). Selection-baseline comparisons use final RL step 64. Diverse and Similar results use step 8 for OpenThoughts3 and Nemotron-Cascade 2 and step 16 for INTELLECT-3. The Dolci diagnostic uses the end of matched SFT, before RL.\\
\addlinespace[6pt]
\rowcolor{AppendixTableHeader}
\multicolumn{2}{@{}l}{\bfseries\color{AppendixTableInk} Single teacher}\\*
Candidates and verification & Qwen3-4B-Thinking-2507 writes every candidate, answering Dolci-Think prompts \citep{olmo3} in mathematics, science, verified synthetic tasks, and instruction following. The answer checker retains 246{,}022 verified rows. Each fingerprint concatenates 137 continuous, 430 tree-traversal, and 64 pattern features, followed by a seed-42 Gaussian projection to 96 dimensions.\\*
Selection conditions & Both conditions select 10{,}000, 25{,}000 and 50{,}000 rows from this one pool, balanced across domains, with at most eight solutions per prompt. Diverse uses MiniBatchKMeans within each domain and farthest-point selection inside each cluster, with budgets in proportion to cluster size. Similar keeps the rows nearest each domain's centroid.\\*
Evaluation & The ten benchmarks are AIME 2025 and 2026, HMMT February 2025, November 2025 and February 2026, Beyond AIME, BRUMO 2025 and 2026, and CMIMC 2025 and 2026. Eight samples per problem at temperature 0.7 and top-$p=0.95$, with a 30{,}720-token completion cap and a 32{,}768-token context, reported as \pass{8}.\\*
Training recipe & Qwen3-4B-Base. SFT runs 300, 700 and 1{,}500 steps for 10{,}000, 25{,}000 and 50{,}000 examples, at learning rate $5\times10^{-6}$, batch 32 and sequences up to 16{,}384 tokens. From each final SFT checkpoint, both conditions receive the same GRPO recipe on one fixed shared prompt set and are evaluated at the same RL step: 50 at 10{,}000 examples and 30 at 25{,}000 and 50{,}000. GRPO uses 128 prompts and 8 responses per step, actor learning rate $10^{-6}$, KL coefficient $0.001$, prompt 4{,}096 and response 12{,}288 tokens, and a binary reward for a correct boxed answer with no learned reward model.\\
\addlinespace[6pt]
\rowcolor{AppendixTableHeader}
\multicolumn{2}{@{}l}{\bfseries\color{AppendixTableInk} Sokoban}\\*
Candidates and verification & Replay-verified solutions to procedurally generated boards, from one shared pool of candidate traces written by single-model and multi-model generators. Fingerprints read the state-action structure of each trace.\\*
Selection conditions & Clustered farthest-point selection for diverse and nearest-centroid selection for similar, from one shared pool (Appendix~\ref{app:selection-method}). 86{,}792 rows per condition.\\*
Evaluation & 500 held-out boards (193 easy, 176 medium, 96 hard, 35 expert), 64 samples per board, reported at \pass{1}, \pass{4} and \pass{64}.\\*
Training recipe & OLMo3-7B, one full SFT epoch at batch 32, learning rate $10^{-5}$ and 16{,}384-token sequences. GRPO for 75 steps on 10{,}591 randomly drawn boards with 128 prompts per step, 8 rollouts, actor learning rate $10^{-6}$, KL coefficient $0.001$, prompt 4{,}096 and response 8{,}192 tokens, and a solved-only reward.\\
\addlinespace[6pt]
\rowcolor{AppendixTableHeader}
\multicolumn{2}{@{}l}{\bfseries\color{AppendixTableInk} Program simulation}\\*
Candidates and verification & Rewrite-system programs adapted from the A::B environment of RLVE, checked by an executable verifier. One model writes every single-model trace, and a roster of open reasoning models writes the multi-model traces (Appendix~\ref{app:rosters}).\\*
Selection conditions & A multi-model corpus against a single-model corpus of equal size, 10{,}000 rows each.\\*
Evaluation & 500 held-out prompts over seven difficulty levels, reported as \pass{1} (mean sampled accuracy), \pass{4} and \pass{64} from 64 samples per prompt.\\*
Training recipe & OLMo3-7B, 300 full-parameter SFT updates at batch 32, learning rate $10^{-5}$ and 16{,}384-token sequences. GRPO for 75 steps on one shared set of RL prompts, with 128 prompts per step, actor learning rate $10^{-6}$, KL coefficient $0.001$, prompt 4{,}096 and response 8{,}192 tokens.\\
\addlinespace[6pt]
\end{longtable}
\arrayrulecolor{black}
\endgroup

\section{An Illustrative Coverage Model}
\label{app:coverage-model}
\label{app:model}

For a fixed prompt $x$, let $M_x^\star$ denote useful reasoning moves and let $C$
denote moves that the policy can sample. Write
$u(C)=|C\cap M_x^\star|/|M_x^\star|$ for their coverage. As an illustrative
approximation, suppose per-rollout success is $p(x)\approx\beta u(C)$, where
$\beta\in(0,1]$ accounts for completing the reasoning after a useful move is
available.

With independent rollouts, the expected coverage of one prompt is
\begin{equation}
  \mathbb{E}[\pass{k}(x)]=1-(1-p(x))^k
  \approx 1-(1-\beta u(C))^k.
  \label{eq:coverage}
\end{equation}
Holding $\beta$ fixed, this expression increases with $u(C)$, and its
derivative with respect to $u$ is $k\beta(1-\beta u)^{k-1}$. Take a diverse and a
similar policy whose shares of the useful moves satisfy $0<u_S<u_D$, with $\beta u_D<1$. Their expected
coverage gap is $(1-\beta u_S)^k-(1-\beta u_D)^k$. It is zero at $k=0$, rises to a
single maximum when treating the sampling budget as continuous, at
$k^\star=\ln\!\big(\ln(1-\beta u_D)/\ln(1-\beta u_S)\big)\big/\ln\!\big((1-\beta u_S)/(1-\beta u_D)\big)$,
and then returns toward zero as both expected coverages approach one.
For integer sampling budgets, the maximum is attained at an adjacent integer.

The same per-rollout success probability sets the mixed-group
probability of Section~\ref{sec:analysis}. For $G>1$ independent rollouts it is
$1-p(x)^G-(1-p(x))^G$, which increases with $p(x)$ while $p(x)<1/2$ and peaks at
$p(x)=1/2$. Raising $u(C)$ on a prompt that the policy solves less than half the
time raises both the expected coverage in Equation~\eqref{eq:coverage}
and the chance of mixed rewards while the resulting success probability remains
below one half. Across prompts,
the share with mixed rewards and the expected coverage at budgets above one both
depend on how success is spread over the prompts. A policy with a slightly lower
mean solve rate can therefore have mixed rewards on more prompts
(Section~\ref{sec:analysis}) when neither policy has a higher success
probability on every prompt. This aggregate comparison allows success
probabilities to cross across prompts, unlike the preceding pointwise example
with $u_D>u_S$.

\section{Additional Related Work}
\label{app:related}

\paragraph{Reward signal in group-relative RL.}
Beyond outcome rewards \citep{deepseekmath,deepseekr1}, methods add
process rewards and finer advantages \citep{prime2025,vineppo2025}, filter
groups whose rewards agree or recover their signal
\citep{dapo2025,noprompt2026}, remove the standard-deviation normalizer
\citep{liu2025drgrpo}, select prompts by reward variance, learnability or
difficulty
\citep{jiang2025vcrl,hu2025vade,qu2025mopps,wu2026hive,parashar2025e2h}, or
keep rollouts diverse
\citep{wang2025forkingtokens,chen2025passktraining,hu2025diver,hu2026uniqueness}.
Analyses find that RL updates touch a small subset of parameters
\citep{mukherjee2025,zhu2025path} and that RL narrows the output distribution
of the policy \citep{yue2025rlboundary,cui2025entropymechanism}. The methods
above
act inside RL, and we act before it.

\Needspace{14\baselineskip}
\section{Limitations and Future Work}
\label{sec:limitations}
A direct test of our account would count distinct route fingerprints
among the $k$ samples per prompt for both policies, before and after RL. The
executable-route diagnostics of Appendix~\ref{sec:olmo-explicit-routes} are the
natural place for this readout, because each route's topology can be read
directly in that setting. Replicating the comparisons across training seeds would add
uncertainty intervals to the reported margins. Extending the mixed-reward
measurement of Section~\ref{sec:analysis} beyond one model and 64 mathematics
prompts, and tracking prompts with mixed rewards through training, would connect
the starting spread of rewards to the later gains. An intervention on reward
availability, for example filtering or reweighting groups so that both conditions
see the same number of prompts with mixed rewards, would test that connection
causally.

Each fingerprint is domain-specific. The OMEGA fingerprint is computed
from annotated strategy steps, and the RLVE fingerprint from lexical cues and the
transitions among them. Validating the RLVE fingerprint against per-route
annotations, breaking the OMEGA capacity sweep down by category, and comparing
the conditions at matched per-prompt success rates would show where the gains
concentrate. Our account predicts that they concentrate on prompts with
intermediate success probabilities.

\end{document}